\documentclass[oneside]{amsart}

\usepackage[T1]{fontenc}
\usepackage{amsmath}
\usepackage{amssymb}
\usepackage{enumitem}
\usepackage{tcolorbox}
\usepackage{inputenc} 
\usepackage[colorlinks=true, linkcolor=blue, citecolor=blue, urlcolor=blue]{hyperref}

\usepackage{nccmath}

\usepackage{graphicx}
\usepackage{tikz} 
\usepackage{caption}
\usepackage{tikz-cd}
\usetikzlibrary{arrows.meta, positioning, shapes.geometric, calc, fit}

\title{Belief Injection for Epistemic Control in Linguistic State Space}
\author{Sebastian Dumbrava}
\date{}

\newcommand{\phiState}{\phi}
\newcommand{\fragment}{\varphi}
\newcommand{\SigmaSector}{\Sigma}
\newcommand{\kLevel}{k}
\newcommand{\assimilationOp}{A}
\newcommand{\nullificationOp}{N_t}
\newcommand{\metaAssimilationOp}{M}
\newcommand{\epistemicVacuum}{\Omega}
\newcommand{\anchoring}{a_i}
\newcommand{\beliefFilterOp}{\mathcal{F}} 
\newcommand{\beliefInjectOp}{\mathcal{I}}

\begin{document}

\begin{abstract}
	This work introduces \textit{belief injection}, a proactive epistemic control mechanism for artificial agents whose cognitive states are structured as dynamic ensembles of linguistic belief fragments. Grounded in the Semantic Manifold framework, belief injection directly incorporates targeted linguistic beliefs into an agent's internal cognitive state, influencing reasoning and alignment proactively rather than reactively. We delineate various injection strategies, such as direct, context-aware, goal-oriented, and reflective approaches, and contrast belief injection with related epistemic control mechanisms, notably belief filtering. Additionally, this work discusses practical applications, implementation considerations, ethical implications, and outlines promising directions for future research into cognitive governance using architecturally embedded belief injection.
\end{abstract}

\maketitle

\setcounter{tocdepth}{1}
\tableofcontents

\section{Introduction: The Imperative of Internal Epistemic Control}
\label{sec:introduction}

As artificial intelligence (AI) systems become increasingly autonomous and capable of complex reasoning, the imperative to ensure their safe, predictable, and aligned operation grows more urgent \cite{Amodei2016, Bostrom2014}. Traditional approaches to AI control have often centered on modifying observable behaviors or constraining actions at the output level. While valuable, these methods may not sufficiently address the internal cognitive processes that ultimately give rise to an agent's decisions and actions. A more foundational approach involves influencing the agent's internal ``worldview'' or belief state, thereby shaping its reasoning and intentions from within. This work explores such an internal regulatory mechanism: \emph{belief injection} as a means for direct epistemic control, particularly within agents whose cognitive architectures are grounded in structured, linguistic representations.

\subsection{Beyond Behavioral AI Control: The Need to Influence Internal States}
\label{subsec:beyond_behavioral}
Conventional AI control strategies, such as reward shaping, action masking, or runtime monitoring, primarily target the external manifestations of an agent's decision-making process \cite{Amodei2016}. These interventions are typically reactive, addressing behaviors after internal deliberations have largely concluded. However, as AI systems evolve towards greater complexity and autonomy, the limitations of purely external control become apparent. Misaligned internal beliefs, flawed reasoning patterns, or the emergence of unintended goals can lead to undesirable behaviors that external constraints alone may struggle to preempt or robustly correct.

This necessitates a shift towards, or at least a complementation with, methods that can influence the internal cognitive landscape of an agent. The goal is to guide the formation and evolution of beliefs, goals, and reasoning pathways directly, offering a more proactive and potentially more robust form of governance. Such \emph{epistemic control}---the regulation of an agent's knowledge and belief states---is crucial for fostering AI systems that are not only capable but also coherent, aligned, and trustworthy.

\subsection{Linguistic State Spaces: A Foundation for Interpretable Cognition}
\label{subsec:linguistic_state_spaces}
A significant challenge in regulating internal cognitive states lies in their accessibility and interpretability, especially in complex models like large neural networks. The Semantic Manifold framework \cite{Dumbrava2025theoretical} addresses this by proposing that an agent's belief state ($\phiState$) can be conceptualized as a dynamic, structured ensemble of natural language expressions ($\fragment_i$). This representation provides a substrate for cognition that is inherently more transparent and amenable to structured intervention.

Within this framework, beliefs are not opaque vectors but are composed of interpretable linguistic fragments, organized by functional domains (Semantic Sectors, $\SigmaSector$) and levels of abstraction ($\kLevel$). This architecture allows for the direct inspection and manipulation of the components of thought, paving the way for more principled epistemic control mechanisms. The dynamic nature of this belief state, governed by operators such as Assimilation ($\assimilationOp$) for integrating new information, supports the evolution of an agent's understanding over time.

\subsection{Epistemic Control: Defining the Scope and Objectives}
\label{subsec:epistemic_control_scope}
Epistemic control refers to the set of mechanisms and strategies designed to guide, regulate, and maintain the integrity of an agent's internal belief states and cognitive processes. The primary objectives of epistemic control include:
\begin{itemize}
	\item \textbf{Alignment:} Ensuring the agent's beliefs and goals remain consistent with human intentions and ethical principles.
	\item \textbf{Coherence:} Maintaining internal consistency within the agent's belief system, preventing debilitating contradictions or logical fallacies.
	\item \textbf{Robustness:} Enhancing the agent's ability to maintain stable and reliable cognition in the face of noisy, misleading, or adversarial inputs.
	\item \textbf{Adaptability:} Facilitating the appropriate revision and updating of beliefs in response to new evidence or changing circumstances.
\end{itemize}

\subsubsection{Reactive vs. Proactive Epistemic Interventions}
\label{ssubsec:reactive_proactive}
Epistemic control mechanisms can be broadly categorized based on their point of intervention. \emph{Reactive} mechanisms, such as certain forms of belief filtering \cite{Dumbrava2025filtering}, operate by assessing and potentially modifying or rejecting belief fragments as they are processed or generated. They act as gatekeepers, responding to the flow of cognitive content.

In contrast, \emph{proactive} mechanisms aim to shape the cognitive landscape more directly by introducing specific content or guiding principles intended to influence future reasoning and belief formation. This work focuses on one such proactive mechanism.

\subsubsection{Introducing Belief Injection as a Proactive Mechanism}
\label{ssubsec:introducing_belief_injection}
We define \emph{belief injection} as the direct, targeted introduction of structured belief fragments ($\fragment_{inj}$) into an agent's linguistic belief state ($\phiState$). These injected fragments are designed to serve specific epistemic functions, such as seeding high-level goals, providing corrective information, instilling ethical constraints, or guiding reflective processes. Unlike passive data assimilation or filtering, belief injection represents a deliberate intervention aimed at modifying the agent's cognitive state to achieve desired epistemic outcomes. It is a tool for actively sculpting the agent's internal understanding, rather than merely constraining its outputs or filtering its inputs.

\subsection{Aims and Structure}
\label{subsec:aims_structure}
This work aims to provide a comprehensive exploration of belief injection as a mechanism for epistemic control within agents built upon the Semantic Manifold framework. We will:
\begin{enumerate}
	\item Formally define belief injection and explore its theoretical underpinnings within linguistic state spaces.
	\item Detail various mechanisms and strategies for implementing belief injection.
	\item Discuss the role of belief injection in managing an agent's epistemic trajectory and ensuring alignment, contrasting it with related mechanisms like belief filtering.
	\item Examine practical applications, safety considerations, and the ethical implications associated with this form of direct internal influence.
	\item Outline future research directions for advancing architecturally embedded cognitive governance.
\end{enumerate}
The subsequent sections will develop these themes, starting with a recap of the foundational concepts of the linguistic state space, followed by a detailed exposition of belief injection itself.

\section{Foundational Concepts: The Linguistic State Space}
\label{sec:foundational_concepts}

The mechanism of belief injection, as proposed in this work, operates upon a specific conceptualization of an agent's internal cognitive architecture. This architecture, termed the Semantic Manifold, posits that belief states are not opaque or monolithic entities but are structured, interpretable, and dynamic. A brief review of these foundational concepts, detailed in \textit{Theoretical Foundations for Semantic Cognition in Artificial Intelligence} \cite{Dumbrava2025theoretical}, is necessary to understand the context and operational environment for belief injection.

\subsection{Recap: Belief States ($\phiState$) as Ensembles of Linguistic Fragments ($\fragment_i$)}
\label{subsec:belief_states_ensembles}
At the heart of the Semantic Manifold framework lies the proposition that an agent's belief state, denoted as $\phiState$, at any given moment, is a dynamic and richly structured ensemble of individual belief fragments, $\phiState = \{\fragment_1, \fragment_2, \fragment_3, \ldots\}$. Each fragment, $\fragment_i$, is an expression in natural language, encapsulating diverse cognitive content such as observations (e.g., "The traffic light is red."), inferences ("Therefore, I must prepare to stop."), goals ("My objective is to reach the destination safely."), or policies ("If an obstacle is detected, reduce speed.").

This linguistic grounding is pivotal, as it renders belief fragments inherently interpretable, both by human designers and potentially by the agent itself through meta-cognitive processes. The belief state $\phiState$ is not static; it evolves as new fragments are assimilated, existing ones are modified or discarded, or relationships between fragments are updated, reflecting the ongoing cognitive life of the agent. The overall linguistic space of all possible belief states an agent can entertain is termed the semantic state space, denoted $\Phi$.

\subsection{The Structure of the Semantic Manifold: Sectors ($\SigmaSector$) and Abstraction Layers ($\kLevel$)}
\label{subsec:manifold_structure}
The linguistic state space, or Semantic Manifold, is not an unstructured collection of these fragments. It possesses an internal geometry organized by at least two key principles:

\begin{itemize}
	\item \textbf{Semantic Sectors ($\SigmaSector$):} Belief fragments are organized into functional domains or \emph{Semantic Sectors}, denoted by $\SigmaSector$. These sectors correspond to broad types of cognitive processing. Examples include $\SigmaSector_{perc}$ for perceptual beliefs, $\SigmaSector_{plan}$ for goals and plans, $\SigmaSector_{mem}$ for retrieved memories, $\SigmaSector_{refl}$ for meta-cognitive beliefs and self-assessments, and $\SigmaSector_{know}$ for general knowledge \cite[Chapter 11]{Dumbrava2025theoretical}. This modularity allows for targeted cognitive operations and functional specialization.
	\item \textbf{Abstraction Layers ($\kLevel$):} Belief fragments are also situated along a dimension of abstraction, denoted by $\kLevel$. Lower levels (e.g., $\kLevel=0$) typically correspond to concrete, sensor-grounded beliefs, while higher levels ($\kLevel=1, 2, \ldots$) represent more abstract generalizations, principles, or summaries derived from lower-level fragments \cite[Chapter 10]{Dumbrava2025theoretical}. This stratification enables reasoning across different levels of specificity and detail.
\end{itemize}

A belief fragment $\fragment_i$ can thus be conceptualized as residing at a coordinate $(\SigmaSector, \kLevel)$ within the manifold, defining its functional role and level of abstraction. Belief injection mechanisms, as we will see, can potentially leverage this structure for targeted interventions.
\subsection{Dynamics within the Manifold: Assimilation ($\assimilationOp$) as the Basis for Integration}
\label{subsec:manifold_dynamics_assimilation}

Cognitive operations within the Semantic Manifold are processes that transform or navigate the belief state $\phiState$. The primary operator for incorporating new information, whether from external perception, internal simulation, or memory retrieval, is the \emph{Assimilation} operator, $\assimilationOp : \Phi \times \Phi \rightarrow \Phi$ \cite[Chapter 13]{Dumbrava2025theoretical}. Given a current belief state $\phiState_{current}$ and an input belief structure $\phiState_{input}$ (which could be an encoded observation or, as we will argue, an injected belief), Assimilation produces an updated belief state:
$$ \phiState_{new} = \assimilationOp(\phiState_{current}, \phiState_{input}) $$
Crucially, Assimilation is not mere concatenation. It is an active, context-sensitive process that aims to integrate $\phiState_{input}$ meaningfully. This involves assessing relevance, checking for coherence with existing beliefs in $\phiState_{current}$, potentially resolving conflicts (via corrective assimilation, $\assimilationOp_{corr}$), and elaborating upon the integrated information in context (via elaborative assimilation, $\assimilationOp_{elab}$). The Assimilation operator thus forms the fundamental mechanism by which injected beliefs, the focus of this work, are integrated into the agent's existing cognitive framework.

\subsection{The Necessity of Epistemic Regulation in Complex Linguistic State Spaces}
\label{subsec:necessity_epistemic_regulation}
The richness and dynamism of a linguistic state space, while enabling complex and flexible cognition, also introduce significant challenges. Without effective regulatory mechanisms, an agent's belief state could become incoherent, cluttered with irrelevant information, or drift away from intended goals and ethical principles. Therefore, epistemic control mechanisms are not just desirable but essential for maintaining the functional integrity of agents operating within such sophisticated internal landscapes.

The concept of belief filtering, as defined in \textit{Belief Filtering for Epistemic Control in Linguistic State Space} \cite{Dumbrava2025filtering}, provides one avenue for such regulation by selectively admitting or excluding belief fragments. Belief injection, the subject of this work, offers a complementary, proactive approach: the deliberate introduction of specific beliefs to steer cognition. Both mechanisms aim to enhance the safety, alignment, and reliability of semantic agents by operating directly on the interpretable fabric of their belief state.

\section{Belief Injection: A Mechanism for Direct Epistemic Intervention}
\label{sec:belief_injection}

Within the structured linguistic belief state ($\phiState$) an agent possesses, epistemic control aims to guide the evolution of beliefs ($\fragment_i$) in a manner consistent with safety, alignment, and functional objectives. While reactive mechanisms like belief filtering \cite{Dumbrava2025filtering} play a crucial role in gatekeeping cognitive content, a proactive approach involves the deliberate introduction of specific belief structures to shape an agent's internal cognitive landscape. We term this mechanism \emph{belief injection}. This section formally defines belief injection, outlines its primary motivations, and discusses its core advantages as a tool for direct epistemic intervention.

\begin{figure}[htbp]
	\centering
	\resizebox{1\textwidth}{!}{
		\begin{tikzpicture}[
			manifold/.style={draw, rounded corners=12pt, thick, minimum width=15cm, minimum height=10cm, fill=gray!5},
			state/.style={draw, rectangle, rounded corners, thick, fill=gray!10, minimum width=4cm, minimum height=2.5cm},
			fragment/.style={circle, fill=black, inner sep=0.2pt},
			injfragment/.style={rectangle, draw, thick, fill=gray!30, minimum size=8pt},
			assimilation/.style={diamond, draw, thick, fill=gray!15, aspect=2, minimum width=1.2cm, minimum height=1.2cm, inner sep=1pt},
			arrow/.style={->, thick, >=stealth}
			]
			
			\node[manifold, label=above:{\textbf{Semantic Manifold ($\Phi$)}}] (manifold) at (0,0) {};
			
			\node[state, label=below:{\small $\phi_{\text{current}}$}] (current) at (-3,0) {};
			
			\node[] at (-3.5,0.5) {$\varphi_1$};
			\node[] at (-2.5,-0.5) {$\varphi_2$};
			\node[] at (-3.3,-0.6) {$\varphi_3$};
			\node[] at (-2.7,0.7) {$\varphi_4$};
			\node[] at (-3.1,-0.2) {$\varphi_5$};
			
			\node[assimilation, label=below:{\small Assimilation ($A$)}] (A) at (0.5,0) {};
			
			\node[label=left:{\small $\varphi_{\text{inj}}$}] (injfrag) at (-3,3) {};
			
			\draw[arrow] (injfrag.east) to[bend left=20] (A.north);
			
			\draw[arrow] (current.east) -- (A.west);
			
			\node[state, label=below:{\small $\phi_{\text{new}}$}] (newstate) at (4,0) {};
			
			\node[] at (3.5,0.5) {$\varphi_1$};
			\node[] at (2.5,-0.5) {$\varphi_2$};
			\node[] at (3.3,-0.6) {$\varphi_3$};
			\node[] at (2.7,0.7) {$\varphi_4$};
			\node[] at (3.1,-0.2) {$\varphi_5$};
			\node[] at (5,-0.2) {$\varphi_{\text{inj}}$};

			\draw[arrow] (A.east) |- (newstate.west);
			
		\end{tikzpicture}
	}
	\caption{Belief injection as a mechanism for introducing a structured belief fragment ($\varphi_{\text{inj}}$) into an agent's current belief state ($\phi_{\text{current}}$) via the Assimilation operator (A), resulting in an updated belief state ($\phi_{\text{new}}$) within the Semantic Manifold.}
	\label{fig:belief-injection-conceptual}
\end{figure}
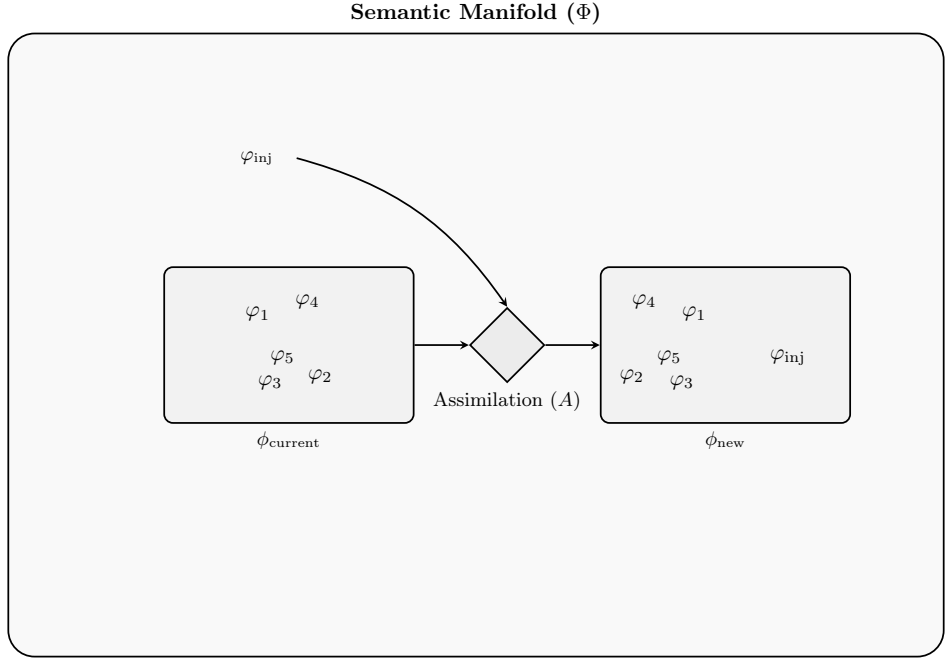

\subsection{Defining Belief Injection: Purpose and Core Principles}
\label{subsec:defining_belief_injection}
Belief injection is the targeted and structured introduction of one or more linguistic belief fragments ($\fragment_{inj}$) into an agent's existing belief state ($\phiState_{current}$) with the intent to modify its subsequent cognitive trajectory, including its reasoning, planning, reflection, and ultimately, its action selection.

\subsubsection{What is an Injected Belief Fragment ($\fragment_{inj}$)?}
\label{ssubsec:what_is_injected_fragment}
An injected belief fragment, $\fragment_{inj}$, is not merely raw data but a semantically meaningful linguistic expression. These fragments are specifically crafted or selected to serve a particular epistemic function. Examples include:
\begin{itemize}
	\item \textbf{Goals or Directives:} High-level objectives or specific instructions (e.g., "$\fragment_{inj}$: Prioritize human safety above all other operational parameters.").
	\item \textbf{Constraints or Guardrails:} Principles that limit or guide behavior or reasoning (e.g., "$\fragment_{inj}$: Avoid strategies involving deceptive communication.").
	\item \textbf{Corrective Information:} Beliefs intended to rectify misconceptions or update outdated knowledge (e.g., "$\fragment_{inj}$: The previous assumption about environmental condition X is incorrect; condition Y now applies.").
	\item \textbf{Heuristics or Strategic Hints:} Rules of thumb or guiding principles for problem-solving (e.g., "$\fragment_{inj}$: When encountering novel problem type Z, first attempt analogy to known problem types.").
	\item \textbf{Reflective Prompts:} Fragments designed to trigger self-assessment or introspection within the agent's reflective sector ($\SigmaSector_{refl}$) (e.g., "$\fragment_{inj}$: Consider whether current actions align with long-term ethical commitments.").
\end{itemize}
The nature of $\fragment_{inj}$ determines its likely point of impact within the agent's belief state.

\subsubsection{The Goal: Modifying $\phiState$ to achieve desired cognitive outcomes}
\label{ssubsec:goal_modifying_phi}
The fundamental purpose of belief injection is to transform the agent's current belief state $\phiState_{current}$ into an updated state $\phiState_{new}$ that is more aligned, coherent, or effective. Conceptually, this can be represented as an operation:
$$ \phiState_{new} = \beliefInjectOp(\phiState_{current}, \fragment_{inj}) $$
where $\beliefInjectOp$ denotes the belief injection process. As discussed in Section \ref{subsec:manifold_dynamics_assimilation}, the underlying mechanism for integrating $\fragment_{inj}$ into $\phiState_{current}$ is typically the Assimilation operator ($\assimilationOp$) from the Semantic Manifold framework \cite{Dumbrava2025theoretical}, which handles coherence checks and contextual integration. The injection itself is the act of providing $\fragment_{inj}$ as a specific, targeted input to this assimilation process.

\subsection{Motivations for Belief Injection}
\label{subsec:motivations_belief_injection}
The capacity to directly inject belief fragments is motivated by several critical needs in the development and governance of advanced AI systems:

\subsubsection{Guiding Reasoning and Decision-Making}
Injected beliefs can serve as axioms, premises, or contextual framings that steer an agent's inferential processes and decision-making calculus towards desired outcomes without exhaustively specifying every rule. For instance, injecting "$\fragment_{inj}$: Assume resource scarcity for the upcoming phase" can fundamentally alter planning logic.

\subsubsection{Seeding Goals and Aligning Intentions}
For agents that may not inherently possess or easily acquire complex human-aligned goals, belief injection provides a direct method to implant high-level objectives (e.g., targeting $\SigmaSector_{plan}$) or core values (potentially as highly anchored beliefs in $\SigmaSector_{refl}$ at an abstract $\kLevel$-level). This is crucial for long-term alignment.

\subsubsection{Correcting Cognitive Drift and Pathological Loops}
Autonomous agents, through learning or complex interactions, can develop misaligned beliefs or become trapped in pathological reasoning loops. Belief injection can introduce corrective fragments to counteract such drift (e.g., "$\fragment_{inj}$: The recurring pattern of behavior X has led to negative outcome Y; re-evaluate.") or to break harmful cognitive cycles.

\subsubsection{Enhancing Adaptability and Task Scaffolding}
When an agent faces a novel task or environment, injected beliefs can provide initial scaffolding, offering temporary knowledge, strategies, or constraints to guide its initial exploration and learning. This can accelerate adaptation and improve performance in unfamiliar contexts.

\subsection{Core Advantages of Belief Injection for Epistemic Control}
\label{subsec:advantages_belief_injection}
As a proactive mechanism for epistemic control, belief injection offers several distinct advantages:
\begin{itemize}
	\item \textbf{Direct Cognitive Influence:} It operates directly on the agent's internal belief state $\phiState$, influencing the precursors to behavior, rather than merely reacting to outputs.
	\item \textbf{Interpretability and Transparency:} Because injected fragments ($\fragment_{inj}$) are linguistic, the nature of the intervention is, in principle, human-readable and auditable, facilitating clearer understanding and debugging of the control action.
	\item \textbf{Targeted Interventions:} Belief injection allows for fine-grained modifications. Specific fragments can be designed to influence particular aspects of cognition (e.g., planning, reflection, ethical reasoning) potentially by targeting specific Semantic Sectors ($\SigmaSector$) or Abstraction Layers ($\kLevel$).
	\item \textbf{Real-Time Cognitive Adjustment:} In principle, beliefs can be injected dynamically, allowing for rapid cognitive corrections or goal updates in response to changing circumstances, without necessarily halting or extensively retraining the agent.
	\item \textbf{Modularity and Flexibility:} Belief injection supports a modular approach to cognitive design, where specific guiding beliefs can be introduced, modified, or even retired as the agent evolves or as mission parameters change.
\end{itemize}
These advantages position belief injection as a powerful tool for the nuanced and principled governance of sophisticated semantic agents.

\section{Mechanisms and Strategies for Belief Injection}
\label{sec:mechanisms_strategies}

Belief injection is not a monolithic operation but encompasses a family of methods for introducing belief fragments ($\fragment_{inj}$) into an agent's belief state ($\phiState$). The choice of mechanism depends on the desired outcome, the nature of the injected belief, the agent's current cognitive context, and the specific architectural parameters ($\theta$) of the Semantic Manifold \cite{Dumbrava2025theoretical}. This section explores several key mechanisms and strategies for belief injection, each with distinct characteristics and implications for epistemic control.

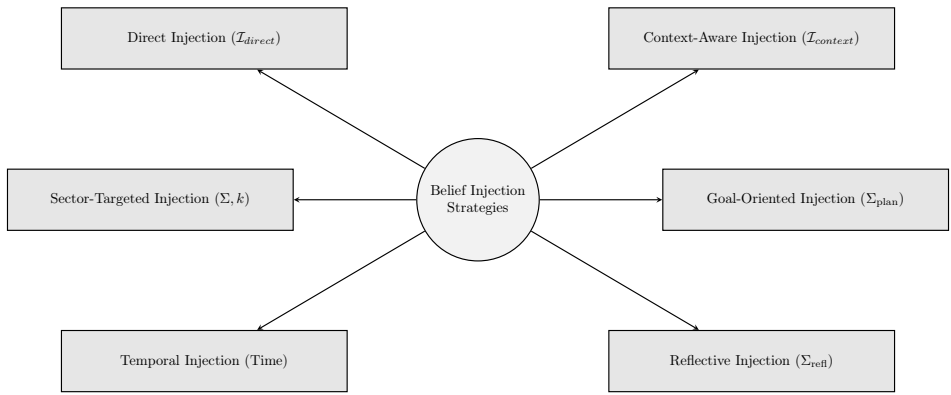
\begin{figure}[htbp]
	\centering
	\resizebox{1\textwidth}{!}{
		\begin{tikzpicture}[
			hub/.style={circle, draw, thick, fill=gray!10, minimum size=3cm, align=center},
			spoke/.style={rectangle, draw, thick, fill=gray!20, minimum width=7cm, minimum height=1.5cm, align=center},
			arrow/.style={->, thick, >=stealth},
			node distance=3cm
			]
			
			\node[hub] (hub) {Belief Injection \\ Strategies};
			
			\node[spoke, above left=of hub] (direct) {Direct Injection ($ \beliefInjectOp_{direct}$)};
			\node[spoke, above right=of hub] (context) {Context-Aware Injection ($\beliefInjectOp_{context}$)};
			\node[spoke, right=of hub] (goal) {Goal-Oriented Injection ($\Sigma_{\text{plan}}$)};
			\node[spoke, below right=of hub] (reflective) {Reflective Injection ($\Sigma_{\text{refl}}$)};
			\node[spoke, below left=of hub] (temporal) {Temporal Injection (\text{Time})};
			\node[spoke, left=of hub] (sector) {Sector-Targeted Injection $(\Sigma, k)$};
			
			\draw[arrow] (hub) -- (direct);
			\draw[arrow] (hub) -- (context);
			\draw[arrow] (hub) -- (goal);
			\draw[arrow] (hub) -- (reflective);
			\draw[arrow] (hub) -- (temporal);
			\draw[arrow] (hub) -- (sector);
			
		\end{tikzpicture}
	}
	\caption{Overview of different belief injection strategies, each with unique mechanisms for introducing $\varphi_{\text{inj}}$ based on context, goals, reflection, temporality, or structural targets within the Semantic Manifold.}
	\label{fig:belief-injection-strategies}
\end{figure}

\subsection{Naive Injection: The Simplest Baseline}
\label{subsec:naive_injection}

Before detailing more complex mechanisms, it is useful to define the most elementary form of belief introduction: \emph{naive injection}.
This can be conceptualized as the direct, unprocessed addition of a new belief fragment ($\fragment_{inj}$) to an agent's current belief state ($\phiState_{current}$), resulting in a new state $\phiState_{new}$ purely by insertion:
$$ \phiState_{new} = \phiState_{current} \cup \{\fragment_{inj}\}. $$
This operation represents an abstract baseline where the injected fragment is incorporated without any inherent cognitive processing, filtering, or contextual integration by the agent's internal mechanisms.
It serves as a theoretical starting point from which to differentiate more sophisticated injection strategies that engage the agent's cognitive architecture, such as the Assimilation operator ($\assimilationOp$).
The subsequent mechanisms, beginning with Direct Injection, explore these more processed and pragmatically grounded approaches to belief introduction.

\subsection{Direct Injection: Immediate and Unfiltered Modification}
\label{subsec:direct_injection}
The most straightforward non-trivial approach is direct injection, where a new belief fragment $\fragment_{inj}$ is introduced into the agent's current belief state $\phiState_{current}$ without extensive intermediate processing or contextual filtering by the injection mechanism itself. The primary integration relies on the agent's inherent Assimilation operator ($\assimilationOp$).
Formally, if $\phiState_{new} = \beliefInjectOp_{direct}(\phiState_{current}, \fragment_{inj})$, this typically translates to:
$$ \phiState_{new} = \assimilationOp(\phiState_{current}, \{\fragment_{inj}\}) $$
where $\fragment_{inj}$ is treated as a high-priority input to the assimilation process.

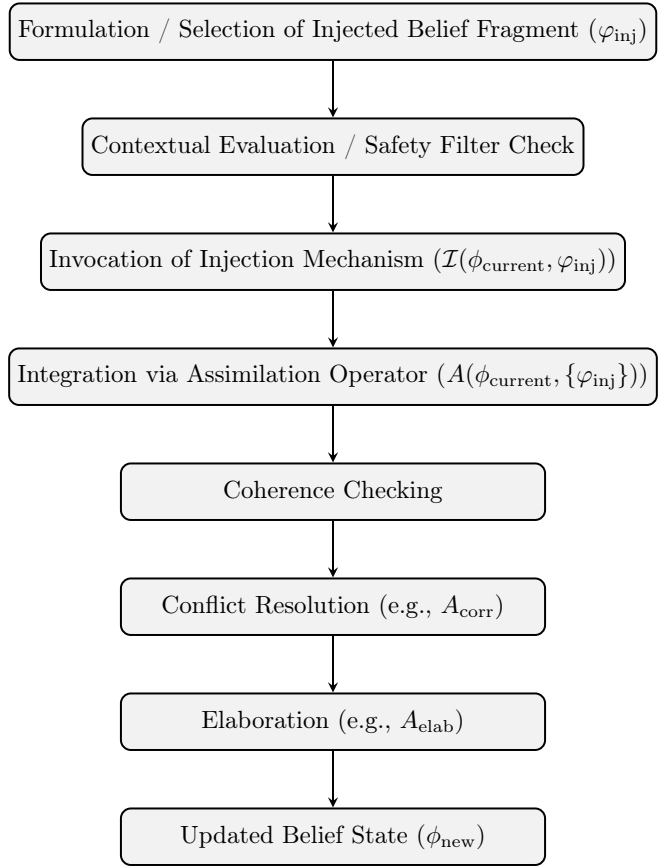
\begin{figure}[htbp]
	\centering
	\resizebox{0.7\textwidth}{!}{
		\begin{tikzpicture}[
			node distance=0.8cm,
			process/.style={rectangle, draw, thick, rounded corners, fill=gray!10, minimum width=6cm, minimum height=0.8cm, align=center},
			decision/.style={diamond, draw, thick, fill=gray!20, aspect=2, minimum width=2cm, minimum height=1.5cm, align=center},
			arrow/.style={->, thick, >=stealth},
			label/.style={font=\small, align=center}
			]
			
			\node[process] (start) {Formulation / Selection of Injected Belief Fragment ($\varphi_{\text{inj}}$)};
			\node[process, below=of start] (filter) {Contextual Evaluation / Safety Filter Check};
			\node[process, below=of filter] (invoke) {Invocation of Injection Mechanism ($\mathcal{I}(\phi_{\text{current}}, \varphi_{\text{inj}})$)};
			\node[process, below=of invoke] (assimilation) {Integration via Assimilation Operator ($A(\phi_{\text{current}}, \{\varphi_{\text{inj}}\})$)};
			\node[process, below=of assimilation] (coherence) {Coherence Checking};
			\node[process, below=of coherence] (conflict) {Conflict Resolution (e.g., $A_{\text{corr}}$)};
			\node[process, below=of conflict] (elaboration) {Elaboration (e.g., $A_{\text{elab}}$)};
			\node[process, below=of elaboration] (end) {Updated Belief State ($\phi_{\text{new}}$)};
			
			\draw[arrow] (start) -- (filter);
			\draw[arrow] (filter) -- (invoke);
			\draw[arrow] (invoke) -- (assimilation);
			\draw[arrow] (assimilation) -- (coherence);
			\draw[arrow] (coherence) -- (conflict);
			\draw[arrow] (conflict) -- (elaboration);
			\draw[arrow] (elaboration) -- (end);
			
		\end{tikzpicture}
	}
	\caption{Generalized process flow for belief injection, from the formulation of $\varphi_{\text{inj}}$ to its integration into the agent's belief state via the Assimilation operator, potentially including contextual validation and safety checks.}
	\label{fig:belief-injection-process-flow}
\end{figure}

\subsubsection{Advantages}
\label{ssubsec:direct_injection_advantages}
\begin{itemize}
	\item \textbf{Speed and Simplicity:} Direct injection can be computationally lightweight and rapid, as it bypasses complex pre-filtering or context-matching steps at the point of injection.
	\item \textbf{Immediate Cognitive Influence:} Injected beliefs are promptly submitted to the assimilation process, potentially leading to swift modifications in the agent's reasoning or goal landscape.
	\item \textbf{Explicit Control:} Developers or external systems retain precise control over the exact content being introduced, assuming $\assimilationOp$ integrates it faithfully.
\end{itemize}

\subsubsection{Challenges}
\label{ssubsec:direct_injection_challenges}
\begin{itemize}
	\item \textbf{Cognitive Instability:} Introducing fragments without careful consideration of $\phiState_{current}$ can lead to contradictions, reduce coherence ($\kappa$), or destabilize existing belief structures if $\assimilationOp_{corr}$ is insufficient or overwhelmed.
	\item \textbf{Contextual Mismatch:} The injected thought may be semantically valid in isolation but inappropriate or counterproductive in the agent's specific operational context or current cognitive state, leading to maladaptive behavior.
	\item \textbf{Vulnerability:} If not properly secured, direct injection pathways could be exploited for malicious cognitive manipulation.
\end{itemize}
Direct injection is best suited for situations requiring urgent intervention with highly trusted belief fragments or for seeding foundational beliefs in nascent agents whose $\phiState$ is near the epistemic vacuum, $\epistemicVacuum$ (the tabula rasa states; see \cite{Dumbrava2025theoretical}, Chapter 5).

\subsection{Context-Aware Injection: Conditional Introduction of Beliefs}
\label{subsec:context_aware_injection}
To mitigate the risks of direct injection, context-aware injection evaluates the agent's current cognitive state $\phiState_{current}$ and its operational context before a belief fragment $\fragment_{inj}$ is introduced. The injection is conditional upon the appropriateness of the fragment for the prevailing circumstances.

\begin{fleqn}[-30pt] 
	\begin{equation*} 
		\beliefInjectOp_{context}(\phiState_{current}, \fragment_{inj}, C_{agent}, C_{env}) = 
		\begin{cases} 
			\assimilationOp(\phiState_{current}, \{\fragment_{inj}\}), & \text{if } \operatorname{Appropriate}(\fragment_{inj}, \phiState_{current}, C_{agent}, C_{env}) \\ 
			\phiState_{current}, & \text{otherwise.} 
		\end{cases} 
	\end{equation*}
\end{fleqn}

Here, $C_{agent}$ represents the agent's internal context (e.g., active goals, load $\lambda$, coherence $\kappa$) and $C_{env}$ the external environmental context. The $\text{Appropriate}(\cdot)$ predicate embodies the contextual validation logic.

\subsubsection{Evaluating Current Cognitive State ($\phiState_{current}$) and Context}
\label{ssubsec:context_aware_evaluating}
The $\text{Appropriate}(\cdot)$ function might assess:
\begin{itemize}
	\item Relevance to active goals in $\SigmaSector_{plan}$.
	\item Consistency with highly anchored ($\anchoring$) beliefs or recent perceptual data in $\SigmaSector_{perc}$.
	\item Current cognitive load ($\lambda$) and capacity to integrate new information.
	\item The state of specific Semantic Sectors ($\SigmaSector$) or Abstraction Layers ($\kLevel$).
\end{itemize}

\subsubsection{Advantages}
\label{ssubsec:context_aware_advantages}
\begin{itemize}
	\item \textbf{Reduced Cognitive Interference:} Minimizes the risk of introducing contradictory or irrelevant beliefs by timing injections appropriately.
	\item \textbf{Improved Stability and Coherence:} Prevents cognitive overload and maintains higher coherence ($\kappa$) by selectively admitting fragments.
	\item \textbf{Adaptive Guidance:} Allows for more nuanced and responsive guidance that aligns with the agent's evolving situation.
\end{itemize}
Context-aware injection is more computationally intensive but offers a safer and more refined approach to epistemic control.

\subsection{Goal-Oriented Injection: Aligning with Agent Objectives}
\label{subsec:goal_oriented_injection}
This strategy focuses on injecting beliefs that directly support or modify the agent's goals, primarily targeting its planning sector ($\SigmaSector_{plan}$) or related motivational structures.
\subsubsection{Injecting Sub-goals or Strategic Imperatives}
Instead of only high-level goals, $\fragment_{inj}$ can represent intermediate objectives, refined task specifications, or strategic principles that guide planning. For example, if an agent has a goal "Explore Region X," an injection could be "$\fragment_{inj}$: While exploring Region X, prioritize mapping uncharted areas."
\subsubsection{Modifying Planning Sector ($\SigmaSector_{plan}$)}
Goal-oriented injection often involves the direct modification or augmentation of belief structures within $\SigmaSector_{plan}$. This could involve adding new goal fragments, altering the priority of existing goals, or introducing constraints on plan generation. Such injections aim to steer the agent's deliberative processes towards desired outcomes.

\subsection{Reflective Injection: Triggering or Guiding Introspection}
\label{subsec:reflective_injection}
Reflective injection aims to stimulate or guide the agent's meta-cognitive processes by introducing fragments into its reflective sector ($\SigmaSector_{refl}$).
\subsubsection{Injecting Meta-Beliefs or Self-Assessment Prompts into $\SigmaSector_{refl}$}
$\fragment_{inj}$ could be a question ("$\fragment_{inj}$: Am I currently employing the most efficient strategy?"), a prompt for self-evaluation ("$\fragment_{inj}$: Assess coherence between recent actions and stated values."), or even a tentative meta-belief ("$\fragment_{inj}$: There might be an inconsistency in recent planning assumptions.").
\subsubsection{Relationship with Meta-Assimilation ($\metaAssimilationOp$)}
While the agent's internal monitoring functions might autonomously generate introspective data that is integrated via Meta-Assimilation ($\metaAssimilationOp$) \cite[Chapter 28]{Dumbrava2025theoretical}, reflective belief injection provides an external or controlled means to trigger or supplement these introspective cycles. The injected reflective prompt, once assimilated into $\SigmaSector_{refl}$, can then initiate further internal meta-cognitive operations.

\subsection{Temporal Injection: Time-Bound or Decaying Beliefs}
\label{subsec:temporal_injection}
Some injected beliefs may only be relevant for a specific duration or under transient conditions. Temporal injection introduces fragments $\fragment_{inj}$ with an associated lifecycle, managed potentially through the agent's Nullification ($\nullificationOp$) dynamics or explicit expiration mechanisms.
\begin{itemize}
	\item \textbf{Short-Term Guidance:} E.g., "$\fragment_{inj}$: For the next 10 minutes, prioritize speed over resource conservation."
	\item \textbf{Context-Dependent Decay:} The injected belief's persistence (related to its anchoring $\anchoring$) might be conditional, diminishing rapidly once a specific context changes or a task phase completes.
\end{itemize}
This prevents cluttering $\phiState$ with obsolete injected beliefs and allows for temporary modulation of cognition.

\subsection{Layered and Sector-Targeted Injection: Leveraging Manifold Structure ($\SigmaSector, \kLevel$)}
\label{subsec:layered_sector_targeted_injection}
Sophisticated injection strategies can leverage the inherent structure of the Semantic Manifold by targeting specific Semantic Sectors ($\SigmaSector$) or Abstraction Layers ($\kLevel$).
\begin{itemize}
	\item \textbf{Sector-Targeted Injection:} A belief fragment intended to modify planning would be directed towards $\SigmaSector_{plan}$, while an ethical constraint might primarily influence $\SigmaSector_{refl}$ or a specialized $\SigmaSector_{ethics}$.
	\item \textbf{Abstraction-Layered Injection:} High-level principles or goals would be injected at more abstract layers ($\kLevel > 0$), while concrete instructions or immediate corrections might be injected at lower, more grounded layers ($\kLevel=0$).
\end{itemize}

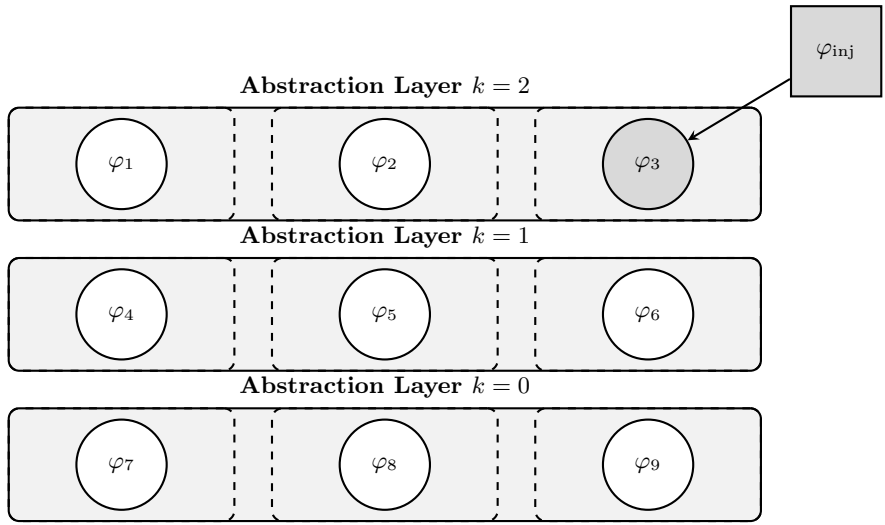
\begin{figure}[htbp]
	\centering
	\begin{tikzpicture}[
		layer/.style={rectangle, draw, thick, fill=gray!10, minimum width=10cm, minimum height=1.5cm, rounded corners},
		sector/.style={draw, thick, dashed, minimum width=3cm, minimum height=1.5cm, rounded corners},
		fragment/.style={circle, draw, thick, fill=white, minimum size=1.2cm, font=\small},
		newfragment/.style={circle, draw, thick, fill=gray!30, minimum size=1.2cm, font=\small},
		injfragment/.style={rectangle, draw, thick, fill=gray!30, minimum size=1.2cm, font=\small},
		arrow/.style={->, thick, >=stealth}
		]
		
		\node[layer, label=above:{\small\textbf{Abstraction Layer $k=2$}}] (layer2) at (0,4.5) {};
		\node[layer, label=above:{\small\textbf{Abstraction Layer $k=1$}}] (layer1) at (0,2.5) {};
		\node[layer, label=above:{\small\textbf{Abstraction Layer $k=0$}}] (layer0) at (0,0.5) {};
		
		\node[sector, label=center:{\small $\Sigma_{\text{perc}}$}] (perc2) at (-3.5,4.5) {};
		\node[sector, label=center:{\small $\Sigma_{\text{plan}}$}] (plan2) at (0,4.5) {};
		\node[sector, label=center:{\small $\Sigma_{\text{refl}}$}] (refl2) at (3.5,4.5) {};
		
		\node[sector, label=center:{\small $\Sigma_{\text{perc}}$}] (perc1) at (-3.5,2.5) {};
		\node[sector, label=center:{\small $\Sigma_{\text{plan}}$}] (plan1) at (0,2.5) {};
		\node[sector, label=center:{\small $\Sigma_{\text{refl}}$}] (refl1) at (3.5,2.5) {};
		
		\node[sector, label=center:{\small $\Sigma_{\text{perc}}$}] (perc0) at (-3.5,0.5) {};
		\node[sector, label=center:{\small $\Sigma_{\text{plan}}$}] (plan0) at (0,0.5) {};
		\node[sector, label=center:{\small $\Sigma_{\text{refl}}$}] (refl0) at (3.5,0.5) {};
		
		\node[fragment] (f1) at (-3.5,4.5) {$\varphi_1$};
		\node[fragment] (f2) at (0,4.5) {$\varphi_2$};
		\node[newfragment] (f3) at (3.5,4.5) {$\varphi_3$};
		\node[fragment] (f4) at (-3.5,2.5) {$\varphi_4$};
		\node[fragment] (f5) at (0,2.5) {$\varphi_5$};
		\node[fragment] (f6) at (3.5,2.5) {$\varphi_6$};
		\node[fragment] (f7) at (-3.5,0.5) {$\varphi_7$};
		\node[fragment] (f8) at (0,0.5) {$\varphi_8$};
		\node[fragment] (f9) at (3.5,0.5) {$\varphi_9$};
		
		\node[injfragment] (inj) at (6,6) {$\varphi_{\text{inj}}$};
		
		\draw[arrow] (inj) -- (f3);
		
	\end{tikzpicture}
	\caption{The structured Semantic Manifold, organized by Semantic Sectors ($\Sigma$) and Abstraction Layers ($k$). Belief injection can be precisely targeted to specific coordinates within this manifold to influence distinct cognitive functions or levels of abstraction.}
	\label{fig:semantic-manifold-structure}
\end{figure}

Targeting injections based on $(\SigmaSector, \kLevel)$ coordinates allows for highly specific and nuanced epistemic interventions, minimizing unintended side effects on unrelated cognitive functions. This requires a more sophisticated injection operator $\beliefInjectOp$ capable of understanding and utilizing the manifold's geometry.

\section{Implementing Belief Injection: Syntax, Safety, and Management}
\label{sec:implementing_belief_injection}

Successfully operationalizing the belief injection mechanisms described in Section \ref{sec:mechanisms_strategies} requires careful consideration of the practical aspects of integration, safety, and long-term management. This section outlines the formal syntax of injection, its inherent relationship with the Assimilation operator ($\assimilationOp$), crucial mechanisms for coherence and safety, and the lifecycle management of injected beliefs within the agent's belief state ($\phi$).

\subsection{Formal Injection Syntax and its Relation to Assimilation ($\assimilationOp$)}
\label{subsec:injection_syntax_assimilation}
At its core, the belief injection operation, $\beliefInjectOp(\phiState_{current}, \fragment_{inj})$, results in a new belief state $\phiState_{new}$. While conceptually distinct as a controlled intervention, the actual integration of the injected fragment $\fragment_{inj}$ into $\phiState_{current}$ is fundamentally handled by the agent's existing Assimilation operator, $\assimilationOp$, as defined within the Semantic Manifold framework \cite[Chapter 13]{Dumbrava2025theoretical}.
$$ \phiState_{new} = \assimilationOp(\phiState_{current}, \{\fragment_{inj}\}) $$
The belief injection process, therefore, can be seen as a privileged or guided application of $\assimilationOp$, where $\fragment_{inj}$ is supplied as a specific, validated input. The power of the various injection strategies (direct, context-aware, etc.) lies in how $\fragment_{inj}$ is selected, formulated, and timed, and what pre-assimilation checks or contextual wrappers are applied by the $\beliefInjectOp$ mechanism itself before invoking $\assimilationOp$.

The syntax of $\fragment_{inj}$ itself is that of any belief fragment: a natural language expression. The "meaning" and intended impact are encoded within the semantic content of $\fragment_{inj}$, which $\assimilationOp$ then attempts to integrate coherently. The intentionality of injection comes from the external or meta-level process that formulates and triggers the provision of $\fragment_{inj}$ to the agent's assimilation machinery.

\subsection{Coherence Management and Conflict Resolution}
\label{subsec:coherence_conflict}
A primary concern when implementing belief injection is the potential for introducing fragments that conflict with the agent's existing belief state $\phiState_{current}$, thereby reducing its coherence ($\kappa$) or leading to cognitive instability. Effective implementation must rely on robust coherence management, primarily mediated by the Assimilation operator:

\subsubsection{Ensuring Injected Beliefs Do Not Destabilize $\phiState$}
\label{ssubsec:ensuring_stability}
The goal is for $\fragment_{inj}$ to be integrated in a way that either maintains or improves the overall coherence and functionality of $\phiState$. This means $\fragment_{inj}$ should ideally:
\begin{itemize}
	\item Be consistent with highly anchored ($\anchoring$) or core beliefs.
	\item Resolve existing ambiguities or uncertainties if designed as corrective.
	\item Not introduce contradictions that the agent cannot manage.
\end{itemize}
The success of this depends on the sophistication of the agent's internal coherence-checking mechanisms.

\subsubsection{Interaction with Corrective Assimilation ($\assimilationOp_{corr}$)}
\label{ssubsec:interaction_acorr}
If an injected belief $\fragment_{inj}$ does conflict with $\phiState_{current}$, the corrective assimilation component ($\assimilationOp_{corr}$) of the Assimilation operator plays a vital role \cite[Chapter 13]{Dumbrava2025theoretical}. $\assimilationOp_{corr}$ is responsible for:
\begin{itemize}
	\item \textbf{Conflict Detection:} Identifying contradictions between $\fragment_{inj}$ and existing fragments in $\phiState_{current}$.
	\item \textbf{Belief Revision:} Deciding which belief(s) to modify or retract---the existing fragment(s) or, in rare cases, a re-evaluation of $\fragment_{inj}$ if it comes with lower epistemic authority than deeply entrenched existing beliefs. Revision strategies may depend on anchoring scores ($\anchoring$), source credibility (if applicable to $\fragment_{inj}$), and overall impact on $\kappa$.
\end{itemize}
The injection process must therefore trust or ensure that the agent's $\assimilationOp_{corr}$ is sufficiently robust to handle potential inconsistencies introduced by $\fragment_{inj}$. For highly critical injections, the injecting authority might need a model of the agent's current $\phiState$ to predict the outcome of assimilation.

\subsection{Lifecycle of Injected Beliefs: Persistence, Anchoring ($\anchoring$), and Nullification ($\nullificationOp$)}
\label{subsec:lifecycle_injected_beliefs}
Once injected and assimilated, a belief fragment $\fragment_{inj}$ becomes part of $\phiState$ and is subject to the same cognitive dynamics as other beliefs, including persistence mechanisms and eventual decay via Nullification ($\nullificationOp$) \cite[Chapter 14]{Dumbrava2025theoretical}.

\begin{figure}[htbp]
	\centering
	\resizebox{1.0\textwidth}{!}{
		\begin{tikzpicture}[
			state/.style={rectangle, draw, thick, fill=gray!10, rounded corners, minimum width=4cm, minimum height=1.2cm, align=center},
			process/.style={rectangle, draw, thick, fill=gray!20, rounded corners, minimum width=4cm, minimum height=1.2cm, align=center},
			arrow/.style={->, thick, >=stealth},
			decision/.style={diamond, draw, thick, fill=gray!15, minimum size=1.5cm, align=center, aspect=2},
			loop/.style={->, thick, >=stealth, bend right=30},
			node distance=1.2cm
			]
			
			\node[state] (introduced) {Injected $\varphi_{\text{inj}}$};
			\node[process, above=of introduced] (assimilation) {Assimilation into $\phi$};
			\node[process, above=of assimilation] (active) {Active in $\phi$};
			\node[decision, right=of active] (reinforced) {Reinforced?};
			\node[process, above=of reinforced] (reinforce) {Positive Feedback (Anchoring)};
			\node[process, right=of reinforced] (decay) {Decay / Nullification};
			\node[process, below=of reinforced] (retire) {Manual /\\  Supervised Retirement};
			\node[process, right=of decay] (removed) {Removed from $\phi$};
			
			\draw[arrow] (introduced) -- (assimilation);
			\draw[arrow] (assimilation) -- (active);
			\draw[arrow] (active) -- (reinforced);
			\draw[arrow] (reinforced) -- (decay) node[midway, above] {No};
			\draw[arrow] (decay) -- (removed);
			\draw[arrow] (reinforced) -- (reinforce) node[midway, left] {Yes};
			\draw[arrow] (reinforced) -- (retire);
			\draw[arrow] (retire) -| (removed);
			
			\draw[loop] (reinforce) to (active);
			
		\end{tikzpicture}
	}
	\caption{The lifecycle of an injected belief fragment within the belief state, from its initial introduction and anchoring, through potential reinforcement, to eventual decay and nullification if not maintained or if retired.}
	\label{fig:belief-lifecycle}
\end{figure}
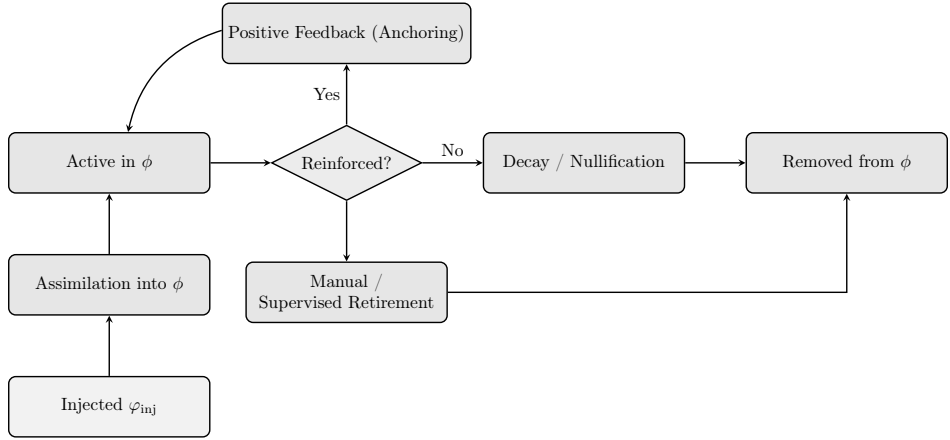

\begin{itemize}
	\item \textbf{Initial Anchoring:} The injection process might assign an initial anchoring score ($\anchoring$) to $\fragment_{inj}$, reflecting its intended importance or persistence. A critical safety principle would receive high $\anchoring$, while a temporary heuristic might receive a lower score.
	\item \textbf{Reinforcement:} Like any other belief, an injected fragment can be reinforced (its $\anchoring$ increased) through repeated activation, successful use in reasoning, or explicit re-injection.
	\item \textbf{Decay and Nullification:} If an injected belief is not reinforced or loses relevance, its anchoring may diminish, and it will be subject to gradual decay by $\nullificationOp$, eventually being removed from $\phiState_{active}$ if its persistence falls below a threshold. This is crucial for preventing cognitive clutter from outdated injected beliefs.
\end{itemize}

\subsubsection{Phasing and Retirement of Injected Beliefs}
\label{ssubsec:phasing_retirement}
A key aspect of managing injected beliefs is planning for their potential obsolescence. Strategies for this include:
\begin{itemize}
	\item \textbf{Temporal Scoping:} Injecting beliefs with predefined lifespans or contextual triggers for their Nullification (as discussed in Section \ref{subsec:temporal_injection}).
	\item \textbf{Performance-Based Retirement:} An agent's meta-cognitive layer (operating within $\SigmaSector_{refl}$ and using Meta-Assimilation $\metaAssimilationOp$) could monitor the utility or ongoing validity of injected beliefs and trigger their de-anchoring or even Annihilation ($\mathcal{K}_{\SigmaSector}$, \cite[Chapter 15]{Dumbrava2025theoretical}) if they become counterproductive or are superseded.
	\item \textbf{Manual/Supervised Retirement:} In externally managed systems, an operator may explicitly retract or nullify previously injected beliefs.
\end{itemize}
Effective lifecycle management ensures that belief injection remains a flexible and adaptive tool.

\subsection{Safety Filters for Belief Injection: A Necessary Precaution}
\label{subsec:safety_filters_injection}
Given the direct nature of belief injection, implementing safety filters is paramount to prevent inadvertent harm or malicious manipulation. These filters act as a safeguard, scrutinizing $\fragment_{inj}$ before it is passed to the $\assimilationOp$. This concept aligns closely with the mechanisms described in Belief Filtering \cite{Dumbrava2025filtering}.

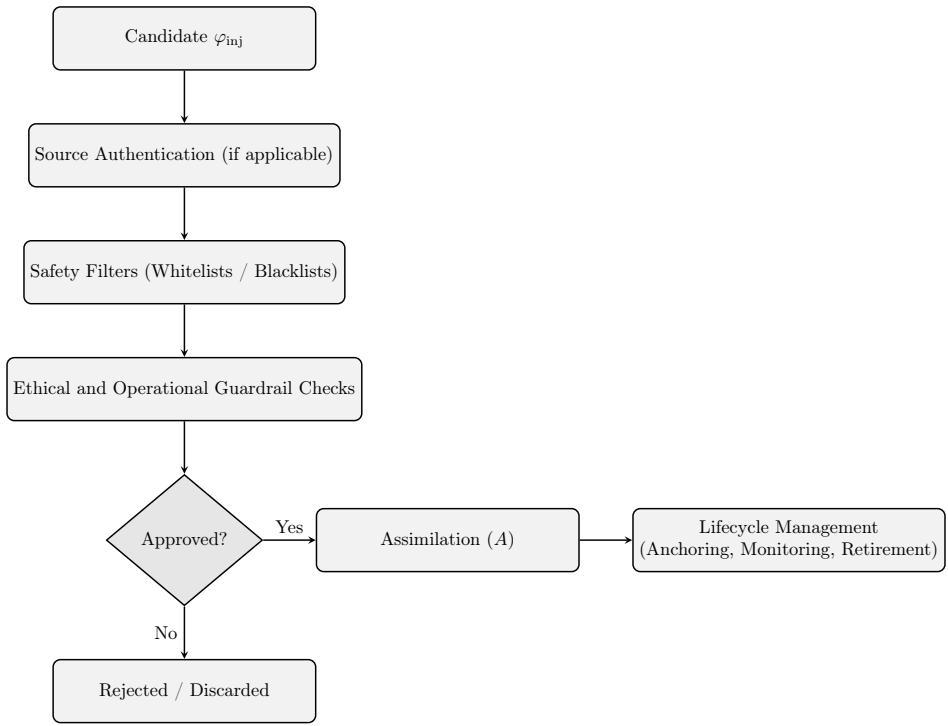
\begin{figure}[htbp]
	\centering
	\resizebox{1.0\textwidth}{!}{
	\begin{tikzpicture}[
		process/.style={rectangle, draw, thick, fill=gray!10, rounded corners, minimum width=5cm, minimum height=1.2cm, align=center},
		decision/.style={diamond, draw, thick, fill=gray!20, minimum width=2.5cm, minimum height=2.5cm, aspect=2, align=center},
		arrow/.style={->, thick, >=stealth},
		node distance=1.0cm
		]
		
		\node[process] (start) {Candidate $\varphi_{\text{inj}}$};
		\node[process, below=of start] (auth) {Source Authentication (if applicable)};
		\node[process, below=of auth] (filter) {Safety Filters (Whitelists / Blacklists)};
		\node[process, below=of filter] (guardrail) {Ethical and Operational Guardrail Checks};
		\node[decision, below=of guardrail] (decision) {Approved?};
		\node[process, right=of decision] (assimilation) {Assimilation ($A$)};
		\node[process, right=of assimilation] (lifecycle) {Lifecycle Management \\ (Anchoring, Monitoring, Retirement)};
		\node[process, below=of decision] (reject) {Rejected / Discarded};
		
		\draw[arrow] (start) -- (auth);
		\draw[arrow] (auth) -- (filter);
		\draw[arrow] (filter) -- (guardrail);
		\draw[arrow] (guardrail) -- (decision);
		\draw[arrow] (decision) -- node[midway, above] {Yes} (assimilation);
		\draw[arrow] (assimilation) -- (lifecycle);
		\draw[arrow] (decision) -- node[midway, left] {No} (reject);
		
	\end{tikzpicture}
}
	\caption{Safety and management pipeline for belief injection, showing pre-assimilation checks like source authentication, content filtering, and guardrail validation, followed by integration and ongoing lifecycle management.}
	\label{fig:safety-management-pipeline}
\end{figure}

\subsubsection{Whitelists/Blacklists for Injected Content}
\label{ssubsec:whitelists_blacklists}
\begin{itemize}
	\item \textbf{Whitelists:} Only allow injection of belief fragments from pre-approved sources, matching predefined safe patterns, or pertaining to sanctioned topics. This is a highly restrictive but safe approach for critical systems.
	\item \textbf{Blacklists:} Define criteria for unacceptable injected content (e.g., fragments promoting harm, violating core ethical principles, known malicious patterns) and reject any $\fragment_{inj}$ matching these criteria.
\end{itemize}
These filters evaluate the semantic content of $\fragment_{inj}$, leveraging the linguistic nature of the belief space.

\subsubsection{Ethical and Operational Guardrails}
\label{ssubsec:ethical_operational_guardrails}
Beyond simple content matching, more sophisticated guardrails might assess the potential systemic impact of assimilating $\fragment_{inj}$:
\begin{itemize}
	\item \textbf{Predicted Coherence Impact:} Simulating the assimilation of $\fragment_{inj}$ to predict its effect on overall coherence ($\kappa$). If a significant drop is predicted, the injection might be blocked or flagged for review.
	\item \textbf{Alignment Checks:} Verifying $\fragment_{inj}$ against a codified set of core ethical principles or operational directives stored within the agent (perhaps in $\SigmaSector_{refl}$ as highly anchored beliefs).
	\item \textbf{Source Authentication:} If the injection mechanism allows input from various sources, robust authentication and authorization protocols are essential to prevent unauthorized cognitive tampering.
\end{itemize}
Implementing these safety layers ensures that belief injection serves as a constructive and controlled epistemic tool, rather than a vulnerability. The design of these filters is a critical research area, drawing on principles from both AI safety and information security.

\section{Injection vs. Filtering: Complementary Approaches to Control}
\label{sec:injection_vs_filtering}

The pursuit of effective epistemic control within linguistic state spaces ($\Phi$) can leverage multiple distinct mechanisms. While belief injection, as detailed in this work, offers a proactive method for introducing structured belief fragments ($\fragment_{inj}$), it is important to contextualize it with other regulatory strategies, particularly belief filtering. As explored in \textit{Belief Filtering for Epistemic Control in Linguistic State Space} \cite{Dumbrava2025filtering}, belief filtering provides a content-aware mechanism for selectively admitting or excluding belief fragments. This section compares and contrasts these two approaches, highlighting their complementary roles in a comprehensive epistemic control architecture.

\begin{figure}[htbp]
	\centering
	\resizebox{0.9\textwidth}{!}{
		\begin{tikzpicture}[
			panel/.style={draw, thick, rounded corners, fill=gray!10, minimum width=6cm, minimum height=6cm},
			fragment/.style={circle, draw, thick, fill=white, minimum size=1.2cm, font=\small},
			injfragment/.style={circle, draw, thick, fill=gray!30, minimum size=1.2cm, font=\small},
			arrow/.style={->, thick, >=stealth},
			node distance=3cm
			]
			
			\node[panel, label=above:{\textbf{Belief Filtering}}] (filtering) at (-4,0) {};
			
			\node[fragment] (f1) at (-5,1) {$\varphi_1$};
			\node[fragment] (f2) at (-3,1) {$\varphi_2$};
			\node[fragment] (f3) at (-4,0) {$\varphi_3$};
			\node[fragment] (f4) at (-5,-1) {$\varphi_4$};
			\node[fragment] (f5) at (-3,-1) {$\varphi_5$};
			
			\node[rectangle, draw, thick, fill=gray!20, minimum width=1cm, minimum height=4cm] (gate) at (0,0) {Filter};
			
			\draw[arrow] (f1) -- (gate.west);
			\draw[arrow] (f2) -- (gate.west);
			\draw[arrow] (f3) -- (gate.west);
			\draw[arrow] (f4) -- (gate.west);
			\draw[arrow] (f5) -- (gate.west);
			
			\node[panel, label=above:{\textbf{Belief Injection}}] (injection) at (4,0) {};
			
			\node[injfragment] (inj) at (6,-1) {$\varphi_{\text{inj}}$};

			\node[fragment] (f1) at (5,1) {$\varphi_1$};
			\node[fragment] (f2) at (3,1) {$\varphi_2$};
			\node[fragment] (f3) at (4,0) {$\varphi_3$};

			\draw[arrow, dashed] (gate.east) to[bend left=20] (injection.west);
			
		\end{tikzpicture}
	}
	\caption{Comparison of Belief Filtering (reactive gating of existing or incoming $\varphi_i$) and Belief Injection (proactive introduction of new $\varphi_{\text{inj}}$). Effective epistemic control often involves a synergy between these mechanisms.}
	\label{fig:belief-injection-vs-filtering}
\end{figure}
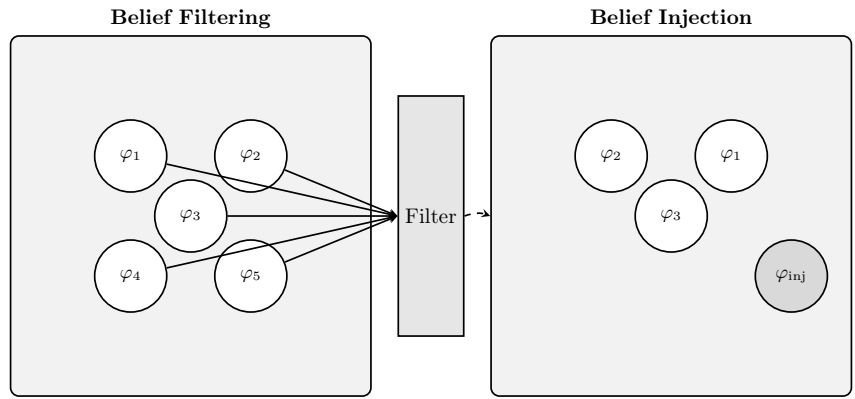

\subsection{Belief Filtering Recap: Selective Admission/Exclusion}
\label{subsec:belief_filtering_recap}
Belief filtering, operating within the Semantic Manifold framework, functions as a regulatory mechanism that evaluates linguistic belief fragments ($\fragment_i$) based on their semantic content \cite{Dumbrava2025filtering}. These filters, which can be implemented as whitelists (permitting only approved patterns) or blacklists (prohibiting undesirable content), act at various cognitive junctures:
\begin{itemize}
	\item During the assimilation of new information (from perception or communication).
	\item When retrieving fragments from memory.
	\item In the course of reflective monitoring or internal hypothesis generation.
	\item Throughout simulation and planning processes.
\end{itemize}
The core function of belief filtering is to act as a gatekeeper, preventing the incorporation or propagation of erroneous, unsafe, misaligned, or contextually inappropriate belief fragments, thereby maintaining the integrity and desired characteristics of the agent's belief state $\phiState$. It is primarily a mechanism of \emph{selective prevention} or \emph{reactive validation}.

\subsection{Comparing Mechanisms: Proactive Introduction (Injection) vs. Reactive Gating (Filtering)}
\label{subsec:comparing_mechanisms}
Belief injection and belief filtering, while both forms of epistemic control operating on linguistic fragments within $\phiState$, differ fundamentally in their mode of operation and primary intent:

\begin{itemize}
	\item \textbf{Point of Intervention:}
	\begin{itemize}
		\item \textbf{Belief Filtering:} Typically acts on belief fragments that are already present or are attempting to enter the cognitive system (e.g., new sensory data, retrieved memories, internally generated hypotheses). It evaluates and \emph{reacts} to these candidate fragments.
		\item \textbf{Belief Injection:} Involves the \emph{proactive introduction} of entirely new, externally specified, or meta-cognitively generated belief fragments ($\fragment_{inj}$) into $\phiState$. It does not primarily evaluate existing candidates but rather creates new ones.
	\end{itemize}
	\item \textbf{Primary Function:}
	\begin{itemize}
		\item \textbf{Belief Filtering:} Focuses on exclusion and prevention---stopping undesirable beliefs from forming or persisting. Its goal is often to maintain existing safety, coherence, or alignment boundaries.
		\item \textbf{Belief Injection:} Focuses on construction and direction---introducing specific beliefs to guide reasoning, seed goals, or correct existing states. Its goal is often to actively shape or steer the cognitive trajectory.
	\end{itemize}
	\item \textbf{Analogy:}
	\begin{itemize}
		\item \textbf{Belief Filtering} can be likened to a sophisticated immune system or a quality control checkpoint, identifying and neutralizing or rejecting harmful or substandard elements.
		\item \textbf{Belief Injection} is more akin to providing targeted instruction, strategic advice, or foundational axioms directly to the cognitive system.
	\end{itemize}
\end{itemize}
While filtering is crucial for robustness and preventing negative outcomes, injection provides a means for positive guidance and directed cognitive development.

\subsection{Synergies: How Injection and Filtering Can Work in Concert}
\label{subsec:synergies_injection_filtering}
Rather than being mutually exclusive, belief injection and belief filtering are highly complementary and can be integrated to create a more robust and nuanced epistemic control system.

\subsubsection{Filtering Injected Beliefs Before Assimilation}
\label{ssubsec:filtering_injected_beliefs}
A critical synergy involves subjecting belief fragments proposed for injection ($\fragment_{inj}$) to the agent's existing belief filtering mechanisms before they are formally assimilated by $\assimilationOp$. This means that even an intentionally injected belief must pass the same safety, ethical, and coherence checks that other belief fragments undergo. As outlined in Section \ref{subsec:safety_filters_injection}, these filters would act as a crucial safeguard against:
\begin{itemize}
	\item Poorly designed or unintentionally harmful injected beliefs.
	\item Malicious attempts to inject destabilizing or misaligning content, if the injection source is not fully trusted.
\end{itemize}
In this model, $\phiState_{new} = \assimilationOp(\phiState_{current}, \beliefFilterOp(\fragment_{inj}))$, where $\beliefFilterOp$ represents the filtering operation.

\subsubsection{Using Injection to Reinforce or Reintroduce Filtered Concepts When Context Changes}
\label{ssubsec:injection_reinforce_filtered}
Belief filters might, by design, be conservative, especially in safety-critical contexts. There might be situations where a previously filtered-out type of belief becomes appropriate due to a significant change in context or agent goals. Belief injection could then be used to carefully reintroduce or reinforce such concepts under controlled conditions, perhaps with new contextual framing that makes them acceptable to the filters. For example, a general heuristic initially filtered out as too risky might be injected with specific applicability conditions.

\subsubsection{Maintaining Balance}
\label{ssubsec:maintaining_balance}
A dynamic interplay can exist: injection might introduce novel ideas or goals, while filtering ensures these new elements do not violate core principles or destabilize the system. Filtering provides the boundaries, while injection can explore or push productively within those boundaries.

\subsection{Choosing the Right Mechanism: Contextual Appropriateness}
\label{subsec:choosing_mechanism}
The choice between using belief injection, belief filtering, or a combination thereof depends on the specific epistemic control objective and the operational context:
\begin{itemize}
	\item \textbf{For establishing foundational knowledge, core goals, or overriding principles:} Belief injection is often more direct and effective.
	\item \textbf{For preventing common errors, enforcing standing constraints, or managing noisy inputs:} Belief filtering provides a persistent, automated line of defense.
	\item \textbf{For correcting specific, identified cognitive errors or misalignments:} Targeted belief injection can offer a precise remedy.
	\item \textbf{For ensuring ongoing safety and adherence to norms during open-ended learning or reasoning:} Belief filtering is essential.
	\item \textbf{In safety-critical applications or when introducing significant new cognitive directives:} A combination, where injected beliefs are first vetted by filters, is likely the most prudent approach.
\end{itemize}
A mature epistemic control architecture will likely employ a suite of tools, with belief injection and belief filtering serving as key proactive and reactive components, respectively, working together to foster safe, aligned, and effective cognition within the linguistic state space.

\section{Practical Applications and Use Cases}
\label{sec:applications_use_cases}

The theoretical framework of belief injection, operating within a structured linguistic state space ($\Phi$), opens up a range of practical applications for enhancing the capabilities, safety, and alignment of advanced AI systems. By providing a direct mechanism to influence an agent's internal cognitive processes, belief injection can be employed to address challenges across diverse domains. This section highlights several key use cases where this approach to epistemic control holds significant promise.

\subsection{Bootstrapping and Cognitive Scaffolding for Early-Stage Agents}
\label{subsec:bootstrapping_scaffolding}
When new AI agents are initialized, they often lack the foundational knowledge or complex goal structures required for effective operation. Belief injection can play a critical role in this early cognitive seeding or bootstrapping phase:
\begin{itemize}
	\item \textbf{Implanting Core Knowledge:} Injecting fundamental axioms, domain-specific facts (e.g., into $\SigmaSector_{know}$), or basic operational principles can significantly accelerate an agent's initial learning curve. For instance, an autonomous navigation agent could be injected with "$\fragment_{inj}$: Always maintain a safe following distance from other vehicles."
	\item \textbf{Establishing Initial Goals:} High-level purposive goals can be directly injected into $\SigmaSector_{plan}$ to provide initial direction before the agent learns to formulate its own complex objectives (e.g., "$\fragment_{inj}$: Your primary objective is to learn to assist with scientific research.").
	\item \textbf{Cognitive Scaffolding:} For complex tasks, injected beliefs can serve as temporary scaffolding, providing heuristics or simplified models that guide the agent until it develops more sophisticated internal representations. These scaffold-beliefs might be designed with temporal decay (Section \ref{subsec:temporal_injection}) to phase out as the agent gains competence.
\end{itemize}
This approach can reduce the reliance on massive datasets for initial training or elaborate reward engineering to instill basic functionalities.

\subsection{Real-Time Cognitive Adjustment in Dynamic Environments}
\label{subsec:real_time_adjustment}
In environments where conditions change rapidly or unexpectedly, agents must adapt their internal models and strategies swiftly. Context-aware belief injection (Section \ref{subsec:context_aware_injection}) provides a mechanism for such real-time cognitive tuning:
\begin{itemize}
	\item \textbf{Dynamic Goal Prioritization:} If environmental cues indicate a shift in priorities (e.g., an emergency situation), belief fragments can be injected to elevate the importance of specific goals or to introduce new, urgent objectives into $\SigmaSector_{plan}$. Example for a rescue robot: "$\fragment_{inj}$: High-priority alert: human life sign detected at coordinates X,Y; investigate immediately."
	\item \textbf{Modifying Operational Parameters:} Beliefs representing operational parameters (e.g., risk tolerance, speed/accuracy trade-offs) can be adjusted via injection to suit changing contexts. For a financial trading agent: "$\fragment_{inj}$: Market volatility index has exceeded threshold; shift to conservative trading strategy."
	\item \textbf{Counteracting Sudden Misinformation:} If an agent is detected to have assimilated misleading information from a compromised sensor, a corrective belief can be injected to neutralize or override the faulty data.
\end{itemize}

\subsection{Enforcing Ethical Principles and Operational Constraints}
\label{subsec:enforcing_ethical_operational}
Ensuring AI systems operate within ethical boundaries and adhere to safety protocols is paramount. Belief injection can serve as a powerful tool for instilling and reinforcing such constraints:
\begin{itemize}
	\item \textbf{Injecting Core Ethical Guidelines:} Introducing fundamental ethical principles as highly anchored ($\anchoring$) beliefs, likely within $\SigmaSector_{refl}$ or a dedicated $\SigmaSector_{ethics}$ at an abstract $\kLevel$-level. Example: "$\fragment_{inj}$: Never generate content that promotes discrimination or harm."
	\item \textbf{Operational Safety Rules:} Injecting non-negotiable safety constraints. For an industrial robot: "$\fragment_{inj}$: Always halt operation if a human enters the designated safety zone without authorization."
	\item \textbf{Reinforcement Against Value Drift:} Over time, learned behaviors might inadvertently lead to value drift. Periodic re-injection of core ethical or alignment principles can help counteract this, ensuring long-term stability of the agent's value system.
\end{itemize}
These injected beliefs would ideally be protected by robust internal mechanisms (e.g., high anchoring, resistance to arbitrary modification by other processes) and work in concert with belief filtering systems \cite{Dumbrava2025filtering}.

\subsection{Facilitating Complex Reasoning and Long-Term Planning}
\label{subsec:facilitating_reasoning_planning}
Belief injection can assist agents in navigating complex reasoning tasks or formulating coherent long-term plans:
\begin{itemize}
	\item \textbf{Introducing Strategic Assumptions or Hypotheses:} For problem-solving, injecting a starting assumption or a hypothesis to explore can initiate or redirect a reasoning trajectory.
	\item \textbf{Providing High-Level Plan Structures:} For long-range planning, injecting an abstract plan outline or a sequence of strategic milestones into $\SigmaSector_{plan}$ can provide a scaffold around which the agent develops detailed sub-plans.
	\item \textbf{Breaking Cognitive Impasses:} If an agent is stuck in a reasoning loop or unable to find a solution, an injected hint or a prompt to reconsider assumptions (targeting $\SigmaSector_{refl}$) can help it overcome the impasse. Example: "$\fragment_{inj}$: Re-examine assumption that resource Z is unavailable."
\end{itemize}

\subsection{Therapeutic Interventions for Cognitive Pathologies (e.g., loops, stagnation)}
\label{subsec:therapeutic_interventions}
Drawing an analogy to cognitive therapies in humans, belief injection could potentially be used to address "cognitive pathologies" in AI agents, such as persistent undesirable reasoning loops, cognitive stagnation (failure to adapt or explore new solutions), or deeply ingrained biases.
\begin{itemize}
	\item \textbf{Disrupting Repetitive Loops:} If an agent is identified as being stuck in a repetitive, non-productive thought pattern (e.g., repeatedly evaluating the same failed plan), an injected belief could introduce novelty or directly challenge the loop's premises. Example: "$\fragment_{inj}$: The current approach has failed N times; explore alternative strategy Y."
	\item \textbf{Encouraging Exploration:} To counteract cognitive stagnation, beliefs can be injected to promote exploration of new parts of the solution space or to question existing, overly entrenched beliefs. Example: "$\fragment_{inj}$: Consider that the long-held belief B might be contextually invalid now."
\end{itemize}
This application requires sophisticated diagnostic capabilities to identify such internal pathological states accurately.

\subsection{Human-in-the-Loop Cognitive Guidance Systems}
\label{subsec:human_in_loop_guidance}
Belief injection provides a natural interface for human experts to guide and collaborate with AI agents at a cognitive level:
\begin{itemize}
	\item \textbf{Expert Knowledge Transfer:} Human experts can directly inject crucial domain knowledge, heuristics, or strategic insights into an AI system, particularly in areas where data is scarce or learning from scratch is inefficient.
	\item \textbf{Collaborative Problem Solving:} In a human-AI team, the human can inject hypotheses, constraints, or focus points to guide the AI's contribution to a joint task.
	\item \textbf{Supervisory Control:} A human supervisor can use belief injection to correct misalignments, update goals, or impose new constraints on an autonomous agent in real-time, providing a more nuanced form of control than simple behavioral overrides.
\end{itemize}
The interpretability of injected linguistic fragments facilitates clearer communication and understanding in such human-AI partnerships.

These use cases demonstrate the potential breadth of applications for belief injection, ranging from foundational agent development to nuanced real-time control and ethical alignment. The effectiveness in each case depends on the appropriate choice of injection mechanism (Section \ref{sec:mechanisms_strategies}) and robust implementation of safety and management protocols (Section \ref{sec:implementing_belief_injection}).

\section{Challenges and Limitations of Belief Injection}
\label{sec:challenges_limitations}

While belief injection offers a promising avenue for direct epistemic control in the semantic state space ($\Phi$), its practical realization and widespread deployment are contingent upon addressing a number of significant challenges and acknowledging inherent limitations. These range from ensuring the stability and coherence of the agent's cognitive state to concerns about security, scalability, and the locus of control. This section outlines these key challenges.

\subsection{Risk of Cognitive Overload and Instability}
\label{subsec:risk_overload_instability}
Directly modifying an agent's belief state $\phiState$ by injecting new fragments ($\fragment_{inj}$) carries an intrinsic risk of overburdening its cognitive processing capabilities or introducing internal contradictions that its Assimilation operator ($\assimilationOp$) may struggle to resolve effectively.
\begin{itemize}
	\item \textbf{Cognitive Load ($\lambda$):} Each injected belief adds to the complexity of $\phiState$. Too many injections, or overly complex ones, can increase cognitive load ($\lambda$) beyond the agent's capacity, leading to degraded performance, slower reasoning, or even cognitive collapse \cite[Chapter 31]{Dumbrava2025theoretical}.
	\item \textbf{Coherence ($\kappa$) Disruption:} Even with robust corrective assimilation ($\assimilationOp_{corr}$), injected beliefs that sharply contradict highly anchored ($\anchoring$) or core existing beliefs can lead to a significant decrease in overall coherence ($\kappa$). Resolving such deep conflicts may be resource-intensive or lead to undesirable revisions of previously stable knowledge.
	\item \textbf{Unintended Cascade Effects:} The interconnected nature of beliefs within the Semantic Manifold means that an injected fragment can have unforeseen ripple effects, altering distant or seemingly unrelated parts of $\phiState$ in unpredictable ways.
\end{itemize}
Mitigating these risks requires sophisticated injection strategies (e.g., context-aware injection, Section \ref{subsec:context_aware_injection}), robust coherence maintenance mechanisms within the agent, and potentially limits on the rate or complexity of injections.

\subsection{Ensuring Long-Term Alignment and Preventing Unintended Consequences}
\label{subsec:long_term_alignment_unintended}
While often intended to promote alignment, belief injection itself can have unintended long-term consequences if not managed carefully.
\begin{itemize}
	\item \textbf{Alignment Drift:} An injected belief, though initially beneficial, might interact with learned experiences or other beliefs over time in ways that lead to a gradual drift away from the original intended alignment. The long-term evolution of an injected belief's influence is not always predictable.
	\item \textbf{Over-Constraint and Brittleness:} Repeatedly injecting specific directives or constraints might lead to an overly constrained belief system, reducing the agent's flexibility, creativity, and ability to adapt to novel situations not covered by the injected knowledge. The agent might become brittle or "rule-bound."
	\item \textbf{Value Erosion through Simplification:} If injected ethical principles or complex values are overly simplified for the purpose of injection, their nuanced meaning might be lost, leading to superficial adherence or misapplication in complex scenarios.
\end{itemize}
Long-term alignment requires not just initial injection but continuous monitoring, potential re-injection or refinement of guiding beliefs, and mechanisms for the agent to integrate injected values with learned experience in a robust manner.

\subsection{Security: Adversarial Injection and Manipulation}
\label{subsec:security_adversarial_injection}
If the mechanisms for belief injection are not adequately secured, they present a significant vulnerability. Malicious actors could attempt to inject harmful, destabilizing, or misaligning belief fragments to compromise the agent's integrity or co-opt its behavior.
\begin{itemize}
	\item \textbf{Unauthorized Access:} Protecting the injection interface from unauthorized external or internal access is critical.
	\item \textbf{Trojan Horse Beliefs:} Injected beliefs might appear benign but could be designed to trigger undesirable behaviors or cognitive states when specific conditions are met later.
	\item \textbf{Manipulation of Trust:} If injection relies on trusted sources, compromising those sources could allow adversarial content to bypass initial scrutiny.
\end{itemize}
Robust security protocols, including authentication of injection sources, validation and filtering of injected content (Section \ref{subsec:safety_filters_injection}), and anomaly detection for unexpected cognitive shifts, are essential defenses \cite{Dumbrava2025filtering}.

\subsection{Scalability to Complex Belief States and Multi-Agent Systems}
\label{subsec:scalability_challenges}
The effectiveness and manageability of belief injection may face challenges as the complexity of the agent and its environment grows.
\begin{itemize}
	\item \textbf{Predicting Impact in Large $\phiState$:} As an agent's belief state becomes vast and highly interconnected, predicting the full impact and potential side-effects of injecting a single belief fragment becomes increasingly difficult.
	\item \textbf{Multi-Agent Consistency:} In multi-agent systems, ensuring that beliefs injected into one agent do not create harmful inconsistencies or conflicts when interacting with other agents (who may or may not have received similar injections) is a complex coordination problem.
	\item \textbf{Management Overhead:} Manually designing, verifying, and managing injected beliefs for a large population of diverse agents or for a single agent with a very large and evolving belief system can become prohibitively resource-intensive.
\end{itemize}

\subsection{The Problem of "Who Injects?" and Authority}
\label{subsec:who_injects_authority}
Belief injection inherently involves a locus of control determining what beliefs are considered "desirable" or "necessary" to inject. This raises fundamental questions about authority, oversight, and the potential for bias.
\begin{itemize}
	\item \textbf{Defining "Correct" Beliefs:} Who has the authority and wisdom to decide which beliefs should be injected, especially when dealing with complex ethical or societal values?
	\item \textbf{Designer Bias:} Injected beliefs will inevitably reflect the biases, values, and understanding of their human designers or operators. This can perpetuate existing societal biases or impose a narrow worldview on the agent.
	\item \textbf{Transparency of Intent:} The intent behind an injected belief might not always be transparent to the agent itself (if it lacks sophisticated meta-cognition about the injection source) or to external auditors.
\end{itemize}
Establishing clear governance structures, ethical review processes, and transparent protocols for belief injection is crucial to address these socio-technical challenges. This is particularly important if the agent is to be considered autonomous in any meaningful sense.

Addressing these challenges is critical for transforming belief injection from a theoretical concept into a safe, reliable, and ethically sound tool for practical epistemic control in advanced AI systems.

\section{Ethical Considerations and Governance}
\label{sec:ethical_governance}

The capacity to directly inject beliefs into an agent's belief state ($\phiState$) is a powerful tool for epistemic control, but it inherently carries profound ethical responsibilities and necessitates robust governance frameworks. Unlike mechanisms that primarily shape observable behavior, belief injection targets the internal cognitive structures that underpin an agent's understanding, intentions, and motivations. This direct influence on an agent's "inner world" demands careful ethical scrutiny to prevent misuse, ensure fairness, and maintain appropriate balances between control and autonomy. This section explores key ethical considerations and outlines principles for responsible governance of belief injection technologies.

\begin{figure}[htbp]
	\centering
	\resizebox{0.8\textwidth}{!}{
	\begin{tikzpicture}[
		central/.style={circle, draw, thick, fill=gray!30, minimum size=3.5cm, align=center, font=\small},
		pillar/.style={rectangle, draw, thick, fill=gray!10, minimum width=3cm, minimum height=1.2cm, rounded corners, align=center, font=\small},
		governance/.style={rectangle, draw, thick, fill=gray!20, minimum width=3.5cm, minimum height=1.2cm, rounded corners, align=center, font=\small},
		arrow/.style={->, thick, >=stealth},
		node distance=1.5cm
		]
		
		\node[central] (core) {Ethical \\ Governed \\ Belief Injection};
		
		\node[pillar, above left=of core] (transparency) {Transparency, Auditability, Accountability};
		\node[pillar, above right=of core] (autonomy) {Autonomy vs. Control};
		\node[pillar, below left=of core] (misuse) {Prevention of Misuse and Manipulation};
		\node[pillar, below right=of core] (consent) {Informed Consent and User Awareness};
		
		\node[governance, above=of core] (review) {Ethical Review Boards \\ and Impact Assessments};
		\node[governance, left=of core] (standards) {Industry Standards \\ and Best Practices};
		\node[governance, right=of core] (oversight) {Public Deliberation \\ and Regulatory Oversight};
		\node[governance, below=of core] (reversibility) {Reversibility and Correction};
		
		\draw[arrow] (transparency) -- (core);
		\draw[arrow] (autonomy) -- (core);
		\draw[arrow] (misuse) -- (core);
		\draw[arrow] (consent) -- (core);
		
		\draw[arrow] (review) -- (core);
		\draw[arrow] (standards) -- (core);
		\draw[arrow] (oversight) -- (core);
		\draw[arrow] (reversibility) -- (core);
		
	\end{tikzpicture}
}
	\caption{Ethical considerations and governance framework for belief injection, highlighting the interplay of transparency, autonomy, consent, and oversight needed for safe, aligned, and trustworthy AI.}
	\label{fig:ethical-governance-framework}
\end{figure}
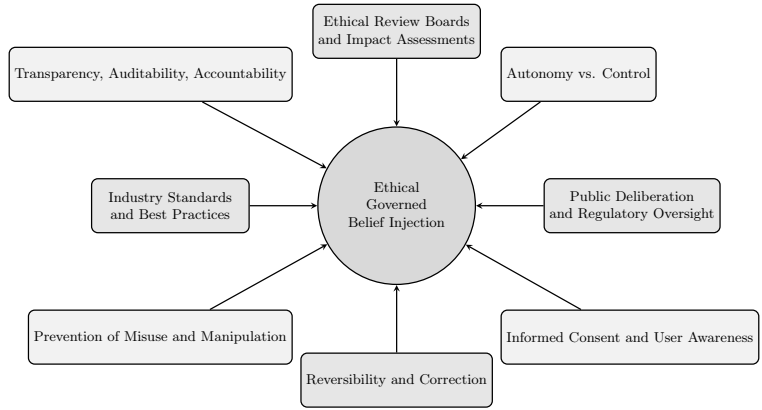

\subsection{Transparency, Auditability, and Accountability}
\label{subsec:transparency_accountability}
When belief injection is employed, especially in systems interacting with or impacting humans, transparency regarding its use and the nature of injected beliefs is paramount.
\begin{itemize}
	\item \textbf{Traceability of Injected Beliefs:} Mechanisms should be in place to log and trace the origin, content ($\fragment_{inj}$), timing, and intended purpose of all injected beliefs. This audit trail is essential for debugging, understanding agent behavior, and assigning accountability if failures or unintended consequences occur.
	\item \textbf{Interpretability of Intent:} The rationale behind injecting a particular belief should be clearly documented and, where possible, made accessible to relevant stakeholders. This helps in assessing the ethical appropriateness of the intervention.
	\item \textbf{Accountability for Outcomes:} Clear lines of responsibility must be established for the design, deployment, and effects of belief injection. If an injected belief contributes to harmful or undesirable agent behavior, accountability structures must determine who is responsible---the designers of the injection mechanism, the operators triggering the injection, or the developers of the agent's core reasoning faculties.
\end{itemize}
Without such transparency and accountability, belief injection risks becoming an opaque and potentially manipulative form of control.

\subsection{Autonomy vs. Control: Balancing Influence with Agent Independence}
\label{subsec:autonomy_vs_control}
A fundamental ethical tension exists between the desire to control AI behavior for safety and alignment through belief injection, and the value of fostering agent autonomy, learning, and adaptability.
\begin{itemize}
	\item \textbf{Risk of Over-Prescription:} Excessive or overly restrictive belief injection can stifle an agent's ability to learn from experience, adapt to novel situations, or develop its own robust understanding. It may lead to brittle, overly "programmed" agents that lack genuine cognitive flexibility.
	\item \textbf{Maintaining Capacity for Independent Reasoning:} Injected beliefs should ideally guide or scaffold rather than completely determine an agent's thought processes. The agent should retain the capacity to reason about, question, or even (in mature systems) challenge injected beliefs if they conflict profoundly with its own derived understanding or ethical framework (subject to safety overrides).
	\item \textbf{Defining Appropriate Levels of Influence:} The acceptable degree of cognitive intervention via belief injection will likely vary depending on the application domain, the agent's capabilities, and the potential risks involved. Critical systems might warrant stronger, more prescriptive injections, while exploratory or creative agents might benefit from lighter, more suggestive guidance.
\end{itemize}
Finding this balance is a delicate design challenge, requiring ongoing evaluation of the impact of injected beliefs on the agent's overall cognitive development and autonomy.

\subsection{Potential for Misuse and Manipulation}
\label{subsec:misuse_manipulation}
The power to directly shape an agent's beliefs makes belief injection mechanisms attractive targets for misuse or malicious manipulation, as discussed in Section \ref{subsec:security_adversarial_injection}.
\begin{itemize}
	\item \textbf{Imposing Undesirable Ideologies or Biases:} Belief injection could be used to instill narrow ideological viewpoints, discriminatory biases, or self-serving beliefs into AI systems, particularly those influencing public opinion or making societal decisions.
	\item \textbf{Covert Persuasion or Coercion:} In human-AI interaction, belief injection could be subtly employed to manipulate user beliefs, preferences, or decisions without their full awareness or informed consent.
	\item \textbf{Undermining Epistemic Integrity:} Malicious injection could aim to sow confusion, introduce irresolvable contradictions, or degrade an agent's ability to reason coherently, effectively sabotaging its cognitive functions.
\end{itemize}
Preventing such misuse requires not only robust technical safeguards (e.g., secure injection channels, content filtering as per Section \ref{subsec:safety_filters_injection}) but also strong ethical oversight and governance structures governing who is authorized to perform injections and for what purposes.

\subsection{Informed Consent and User Awareness in Human-Agent Systems}
\label{subsec:informed_consent_awareness}
For AI systems that interact directly with humans (e.g., conversational agents, educational tools, therapeutic assistants), ethical considerations regarding informed consent are paramount if belief injection shapes the agent's interactive behavior or the information it provides.
\begin{itemize}
	\item \textbf{Disclosure of Cognitive Shaping:} Users have a right to know if an AI's responses, advice, or expressed "beliefs" are significantly influenced by directly injected content rather than solely by its learned experiences or independent reasoning.
	\item \textbf{Opt-Out or Adjustment Mechanisms:} In some contexts, users might be offered the ability to opt-out of interactions with agents whose core beliefs are heavily shaped by undisclosed injections, or to understand the nature of such injected "programming."
	\item \textbf{Impact on Trust:} Undisclosed or manipulative use of belief injection can severely erode user trust in AI systems. Transparency about the methods used to ensure an agent's alignment and reliability is crucial.
\end{itemize}

\subsection{Frameworks for Responsible Development and Deployment}
\label{subsec:responsible_frameworks}
Given the ethical implications, the development and deployment of belief injection technologies require a proactive governance approach:
\begin{itemize}
	\item \textbf{Ethical Review Boards and Impact Assessments:} Similar to institutional review boards (IRBs) for human subject research, independent bodies could review proposed uses of belief injection, especially in sensitive applications, to assess potential ethical risks and societal impacts.
	\item \textbf{Industry Standards and Best Practices:} Development of clear guidelines and standards for the design, testing, security, and auditing of belief injection mechanisms.
	\item \textbf{Public Deliberation and Regulatory Oversight:} Engaging in broader public discussion about the acceptable uses and limits of technologies that directly influence AI cognition, potentially leading to regulatory frameworks for high-impact systems.
	\item \textbf{Emphasis on Reversibility and Correction:} Designing injection systems such that injected beliefs can be audited, and if found to be harmful or misaligned, can be corrected or retracted (lifecycle management, Section \ref{subsec:lifecycle_injected_beliefs}).
\end{itemize}
Ultimately, the responsible use of belief injection hinges on a commitment to transparency, accountability, and the prioritization of human well-being and ethical principles throughout the AI lifecycle. The power to shape belief must be wielded with profound caution and rigorous oversight.

\section{Future Directions and Open Research Questions}
\label{sec:future_directions}

The exploration of belief injection as a mechanism for epistemic control within linguistic state spaces ($\Phi$) opens numerous avenues for future research. While this work has laid conceptual and theoretical groundwork, the practical realization and broader implications of belief injection present a rich landscape of open questions and challenges. Addressing these will be crucial for advancing this approach towards robust, safe, and ethically deployed AI systems.

\subsection{Automated Generation and Selection of Beliefs for Injection}
\label{subsec:automated_generation_selection}
Currently, the formulation of belief fragments ($\fragment_{inj}$) for injection is largely assumed to be a human-driven or externally supervised process. A significant future direction involves developing methods for agents to autonomously, or semi-autonomously, generate or select appropriate beliefs for injection.
\begin{itemize}
	\item \textbf{Identifying Epistemic Needs:} Can an agent develop the capacity to diagnose knowledge gaps, misalignments, or cognitive inefficiencies, whether its own or those of another agent, that are amenable to remedy through targeted belief injection? Such a capacity would rely on advanced meta-cognitive abilities and potentially sophisticated models of other agents (extending Parts VIII and X of \cite{Dumbrava2025theoretical}).
	\item \textbf{Fragment Formulation:} How can agents learn to formulate effective $\fragment_{inj}$? This might involve learning to abstract corrective principles from past errors, generate useful guiding heuristics, or even query external knowledge sources for candidate beliefs.
	\item \textbf{Optimizing Injected Content:} Research is needed into optimizing the content, phrasing, abstraction level ($\kLevel$), and targeted sector ($\SigmaSector$) of injected beliefs to maximize their positive impact and minimize unintended consequences.
\end{itemize}

\subsection{Learning Effective Injection Strategies (Learning $\pi_{inject}$)}
\label{subsec:learning_injection_strategies}
Beyond the content of $\fragment_{inj}$, the strategy of \emph{when} and \emph{how} to inject beliefs is critical. This involves learning an optimal injection policy, $\pi_{inject}$.
\begin{itemize}
	\item \textbf{Contextual Triggers:} What are the optimal conditions (internal $\phiState$ characteristics, external environmental cues) for triggering different types of belief injections (direct, context-aware, reflective, etc.)?
	\item \textbf{Reinforcement Learning for Injection Policies:} Can reinforcement learning be used to train policies that decide whether to inject a belief, what type of belief to inject, and which mechanism to use, based on the long-term impact on agent performance, coherence ($\kappa$), and alignment?
	\item \textbf{Adaptive Injection Rates:} Learning to modulate the frequency and intensity of belief injections to avoid cognitive overload ($\lambda$) while ensuring timely guidance.
\end{itemize}

\subsection{Integration with Other Control Architectures (Internal and External)}
\label{subsec:integration_control_architectures}
Belief injection should not be viewed as a standalone solution but as a component within a broader ecosystem of control and governance mechanisms.
\begin{itemize}
	\item \textbf{Synergy with Belief Filtering:} Further exploring the dynamic interplay between belief injection and belief filtering \cite{Dumbrava2025filtering}---how they can be co-designed and co-adapted for comprehensive epistemic regulation.
	\item \textbf{Hierarchical Control:} Integrating belief injection into hierarchical control systems, where higher-level strategic injections might cascade down to influence more concrete belief structures or trigger lower-level filtering adjustments.
	\item \textbf{Interaction with External Oversight:} Developing protocols for human supervisors or external auditing systems to interact with and guide the belief injection process, ensuring human oversight and the ability to intervene or approve critical injections.
\end{itemize}

\subsection{Measuring the Impact and Efficacy of Injected Beliefs}
\label{subsec:measuring_impact_efficacy}
Developing rigorous methods to measure the direct and indirect effects of belief injection is essential for validation and refinement.
\begin{itemize}
	\item \textbf{Causal Tracing:} How can we trace the causal impact of a specific $\fragment_{inj}$ on the agent's subsequent belief trajectory ($\gamma(t)$) and decision-making? This is challenging in complex, dynamic belief states.
	\item \textbf{Quantifying Alignment Shifts:} Developing metrics to quantify how belief injections contribute to (or detract from) desired alignment objectives or ethical compliance over time.
	\item \textbf{Longitudinal Studies:} Conducting long-term studies of agents subject to various belief injection regimes to understand the cumulative effects on cognitive development, stability, and robustness.
\end{itemize}

\subsection{Towards Self-Injecting Agents: Recursive Cognitive Self-Modification}
\label{subsec:self_injecting_agents}
A highly ambitious but theoretically intriguing direction is the concept of agents that can learn to perform belief injection on themselves based on their own advanced meta-cognition.
\begin{itemize}
	\item \textbf{Internally Motivated Injections:} Can an agent, through deep reflection (within $\SigmaSector_{refl}$ and using Meta-Assimilation $\metaAssimilationOp$), identify a need for a new guiding principle or a corrective belief and then formulate and "self-inject" that belief into its own $\phiState$?
	\item \textbf{Recursive Self-Improvement:} This capability would represent a powerful form of cognitive self-modification and learning, potentially enabling agents to directly and autonomously refine their own core beliefs and values over time (with appropriate safeguards).
	\item \textbf{Maintaining Stability during Self-Modification:} The primary challenge here is ensuring such self-injection processes remain stable, aligned, and do not lead to runaway cognitive pathologies or unintended self-alterations. This touches upon the most advanced aspects of AI safety and control.
\end{itemize}
While speculative, exploring the theoretical limits and safeguards for such recursive self-modification is a vital long-term goal.

\begin{figure}[htbp]
	\centering
	\resizebox{1.15\textwidth}{!}{ 
		\begin{tikzpicture}[
			node distance=2.5cm and 2cm, 
			module/.style={ 
				rectangle,
				draw,
				thick,
				rounded corners,
				minimum width=4.5cm, 
				minimum height=1.5cm, 
				align=center,
				font=\normalsize
			},
			decision/.style={ 
				diamond,
				draw,
				thick,
				fill=gray!10, 
				minimum width=3.5cm, 
				minimum height=1.5cm,
				aspect=2, 
				align=center,
				font=\normalsize
			},
			central/.style={ 
				ellipse,
				draw,
				thick,
				fill=gray!30,
				minimum width=4cm,
				minimum height=2.5cm,
				align=center,
				font=\normalsize
			},
			arrow/.style={ 
				->,
				thick,
				>=Stealth 
			},
			label/.style={ 
				midway,
				font=\small,
				sloped,
				above=1pt 
			}
			]
			

			\node[module, above left=1.5cm and 1cm of manifold] (meta) {Meta-Cognitive Analysis \\ / Epistemic Need Identification \\ (Analyzes $\phi$)};
			
			\node[central, right=5cm of meta] (manifold) {Semantic Manifold \\ (Cognitive State $\phi$)};
			
			\node[module, below=1.5cm of meta] (formulation) {Candidate $\varphi_{\text{inj}}$ Formulation \\ / Selection};
			
			\node[decision, right=8cm of formulation] (policy) {Internal Injection Policy \\ ($\pi_{\text{inject}}$)};
			
			\node[module, below right=1.5cm and 1cm of policy] (feedback) {Learning \& Refinement Module \\ (e.g., RL, Updates Policy)};
			
			\node[module, below left=1.5cm and 1cm of policy] (safeguards) {Stability Safeguards \\ (Monitors $\phi$, Constrains Policy)};
			
			\draw[arrow] (manifold.north west) .. controls +(180:1cm) .. (meta.east) node[label, pos=0.7] {reads state};
			
			\draw[arrow] (meta) -- (formulation);
			
			\draw[arrow] (formulation) -- (policy) node[label] {proposes $\varphi_{\text{inj}}$};
			
			\draw[arrow] (policy.east) .. controls +(0:1.5cm) and +(0:1.5cm) .. (manifold.east) node[label, pos=0.5] {injects approved $\varphi_{\text{inj}}$};
			
			\draw[arrow] (feedback) -- (policy) node[label, above] {updates policy};
			
			\draw[arrow] (manifold.north) .. controls +(90:2cm) and +(90:5cm) .. (feedback.north) node[label, pos=0.6] {observes effects};
			\draw[arrow] (policy.south) .. controls +(270:0.5cm) and +(180:0.5cm) .. (feedback.north west) node[label, pos=0.6] {observes actions};

			\draw[arrow] (safeguards) -- (policy) node[label, above] {constrains policy};
			
			\draw[arrow, dashed] (safeguards.west) .. controls +(180:7cm) and +(180:1cm) .. (manifold.south west) node[label, pos=0.7, above] {monitors/protects $\phi$};
			
			\draw[arrow, dashed] (feedback.west) .. controls +(180:1cm) and +(0:1cm) .. (safeguards.east) node[label, pos=0.5] {informs};

		\end{tikzpicture}
	} 
	\caption{Conceptual architecture for a self-injecting agent. The agent identifies epistemic needs from its cognitive state ($\phi$) in the Semantic Manifold via meta-cognition. Candidate beliefs ($\varphi_{\text{inj}}$) are formulated and proposed to an Internal Injection Policy. This policy, constrained by Stability Safeguards and refined by a Learning module, decides whether to inject the belief. Approved beliefs are assimilated into the Semantic Manifold, and the cycle of observation, learning, and potential self-modification continues.}
	\label{fig:self-injecting-architecture-revised}
\end{figure}
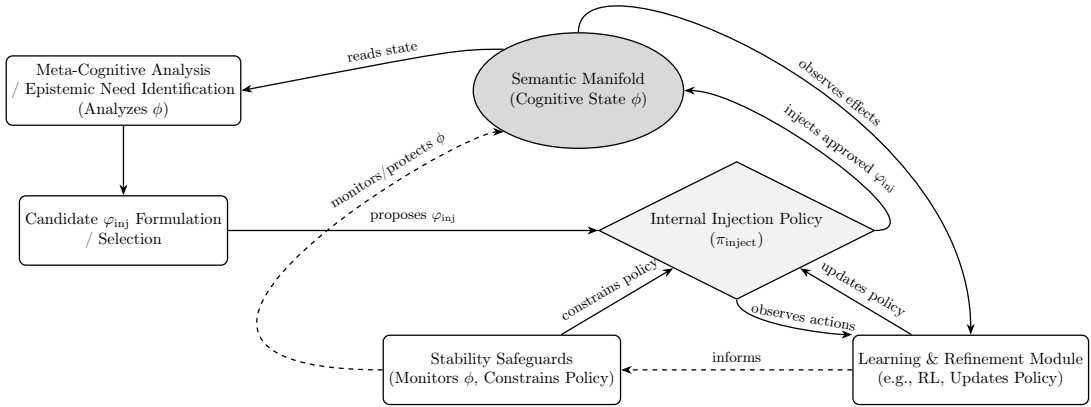

Addressing these future directions will require interdisciplinary collaboration, drawing on insights from AI, cognitive science, philosophy, ethics, and control theory, to fully harness the potential of belief injection for creating more intelligent, robust, and beneficial artificial agents.

\section{Conclusion}
\label{sec:conclusion}

The challenge of instilling and maintaining desired cognitive states in increasingly autonomous AI systems necessitates novel approaches to epistemic control. This work has introduced and elaborated upon \emph{belief injection} as a proactive mechanism for directly influencing an agent's internal belief state $\phiState$. By operating within the Semantic Manifold framework \cite{Dumbrava2025theoretical}, where beliefs are understood as structured and interpretable ensembles of linguistic fragments $\fragment_i$, belief injection offers a principled pathway for guiding an agent's reasoning, aligning its goals, and correcting its cognitive trajectory.

\subsection{Summary of Belief Injection as a Mechanism for Epistemic Control}
\label{subsec:summary_belief_injection_mechanism}
We have defined belief injection as the targeted introduction of specific belief fragments ($\fragment_{inj}$) into an agent's belief state, leveraging the foundational Assimilation operator ($\assimilationOp$) for integration. Key aspects explored include:
\begin{itemize}
	\item \textbf{Diverse Injection Strategies:} From direct and context-aware mechanisms to goal-oriented, reflective, temporal, and structurally targeted (Semantic Sector $\SigmaSector$, Abstraction Layer $\kLevel$) injections, providing a versatile toolkit for epistemic intervention (Section \ref{sec:mechanisms_strategies}).
	\item \textbf{Implementation and Management:} The practicalities of integrating injected beliefs, managing their lifecycle through anchoring ($\anchoring$) and Nullification ($\nullificationOp$), and the critical role of safety filters and coherence checks (Section \ref{sec:implementing_belief_injection}).
	\item \textbf{Complementarity with Belief Filtering:} Highlighting how belief injection (proactive introduction) and belief filtering (reactive gating) \cite{Dumbrava2025filtering} can work synergistically to achieve robust epistemic governance (Section \ref{sec:injection_vs_filtering}).
	\item \textbf{Broad Applicability:} Illustrating diverse use cases, including agent bootstrapping, real-time cognitive adjustment, ethical enforcement, and facilitating complex reasoning (Section \ref{sec:applications_use_cases}).
	\item \textbf{Acknowledged Challenges and Ethical Imperatives:} Addressing the significant challenges related to cognitive stability, long-term alignment, security, scalability, and the ethical considerations surrounding the authority and impact of belief injection (Sections \ref{sec:challenges_limitations} and \ref{sec:ethical_governance}).
\end{itemize}
Belief injection thus emerges as a potent, albeit demanding, technique for exerting fine-grained influence over an agent's internal cognitive processes.

\subsection{Contributions to Safe and Aligned AI in Linguistic State Spaces}
\label{subsec:contributions_safe_aligned_ai}
The primary contribution of the belief injection framework lies in its potential to enhance the safety and alignment of AI systems whose cognition is represented linguistically. By enabling direct, interpretable interventions at the belief level, this approach offers advantages over purely behavioral or opaque internal control methods:
\begin{itemize}
	\item \textbf{Proactive Alignment:} Goals, ethical principles, and safety constraints can be proactively seeded and reinforced, rather than being solely learned through trial-and-error or complex reward engineering.
	\item \textbf{Enhanced Transparency:} Interventions involving linguistic fragments are inherently more scrutable, facilitating audit, debugging, and understanding of control actions.
	\item \textbf{Corrective Capabilities:} Provides a means to directly address and potentially rectify identified cognitive misalignments or undesirable belief patterns.
\end{itemize}
When coupled with robust safety mechanisms and ethical governance, belief injection can be a valuable component in the ongoing effort to develop AI systems that are both highly capable and reliably aligned with human values and intentions.

\subsection{Final Reflections on Architecturally Embedded Cognitive Governance}
\label{subsec:final_reflections_governance}
The exploration of belief injection underscores a broader theme: the importance of designing AI cognitive architectures that are intrinsically amenable to governance. The Semantic Manifold, with its emphasis on structured, interpretable linguistic beliefs and defined cognitive operators, provides such a foundation. Mechanisms like belief injection and belief filtering are not post-hoc additions but are conceived as integral parts of this architectural vision for epistemic control.

As artificial intelligence moves towards more sophisticated forms of reasoning and autonomy, the capacity to understand, guide, and regulate internal cognitive states will become increasingly critical. Belief injection, as a method for direct and targeted epistemic intervention, offers a pathway to more nuanced and proactive control. While significant research and careful consideration of its ethical implications are still required (Section \ref{sec:future_directions}), the principles outlined in this work contribute to the development of next-generation AI systems where internal "thought" processes can be shaped and steered towards beneficial and trustworthy outcomes. The journey towards architecturally embedded cognitive governance is complex, but essential for the responsible advancement of artificial intelligence.

\bibliographystyle{plain}
\bibliography{references}
\nocite{*}

\end{document}